\documentclass[preprint,review,12pt]{elsarticle}

\usepackage{amsmath,amssymb,amsfonts}
\usepackage{graphicx}
\usepackage[graphicx]{realboxes}
\usepackage{textcomp}

\usepackage{xcolor}

\usepackage{subcaption}
\usepackage{hyperref}
\usepackage{dblfloatfix} 

\usepackage[export]{adjustbox}
\usepackage{siunitx}
\usepackage{gensymb}
\usepackage{interval}

\usepackage{enumitem}
\usepackage{dsfont}
\usepackage{multirow}

\usepackage{footnote}
\makesavenoteenv{tabular}

\usepackage{booktabs}
\makeatletter
\newcommand\footnoteref[1]{\protected@xdef\@thefnmark{\ref{#1}}\@footnotemark}
\makeatother

\renewcommand{\color}[2]{#2}

\DeclareMathOperator*{\argmin}{arg\,min}

\begin{document}

\begin{frontmatter}

\title{An automatic framework for Fusing Information from differently stained consecutive digital whole slide images: A case study in renal histology}

\author[add1,add2]{Odyssee Merveille\corref{cor1}}
\author[add1]{Thomas Lampert}
\author[add3]{Jessica Schmitz}
\author[add5]{Germain Forestier}
\author[add3,add4]{Friedrich Feuerhake}
\author[add1]{C\'{e}dric Wemmert}

\cortext[cor1]{odyssee.merveille@creatis.insa-lyon.fr}

\address[add1]{ICube, University of Strasbourg, CNRS (UMR 7357), Strasbourg, France}
\address[add2]{Univ Lyon, INSA-Lyon, Universit\'e Claude Bernard Lyon 1, UJM-Saint Etienne, CNRS, Inserm, CREATIS UMR 5220, U1206, F‐69XXX, LYON, France}
\address[add3]{Institute of Pathology, Hannover Medical School, Germany}
\address[add4]{University Clinic, Freiburg, Germany}
\address[add5]{IRIMAS, Universit\'{e} de Haute Alsace, Mulhouse, France}

\begin{abstract}
\emph{Objective:} This article presents an automatic image processing framework to extract quantitative high-level information describing the micro-environment of glomeruli in consecutive whole slide images (WSIs) processed with different staining modalities of patients with chronic kidney rejection after kidney transplantation.

\emph{Methods:} This four-step framework consists of: 1) approximate rigid registration, 2) cell and anatomical structure segmentation 3) fusion of information from different stainings using a newly developed registration algorithm 4) feature extraction.

\emph{Results:} Each step of the framework is validated independently both quantitatively and qualitatively by pathologists. An illustration of the different types of features that can be extracted is presented.

\emph{Conclusion:} The proposed generic framework allows for the analysis of the micro-environment surrounding large structures that can be segmented (either manually or automatically). It is independent of the segmentation approach and is therefore applicable to a variety of biomedical research questions.

\emph{Significance:} Chronic tissue remodelling processes after kidney transplantation can result in interstitial fibrosis and tubular atrophy (IFTA) and glomerulosclerosis. This pipeline provides tools to quantitatively analyse, in the same spatial context, information from different consecutive WSIs and help researchers understand the complex underlying mechanisms leading to IFTA and glomerulosclerosis.
\end{abstract}


\end{frontmatter}

\section{Introduction}
\label{Sec: Introduction}

More than $90\,000$ kidney transplants are performed each year\footnote{\href{http://www.transplant-observatory.org/download/2017-activity-data-report}{www.transplant-observatory.org/download/2017-activity-data-report}}. Kidney replacement therapy after renal failure can restore renal function for many years, thereby reducing the burden for individual patients and for health systems that are associated with hemodialyis. In the past decades, successful therapy strategies were developed to avoid acute rejection, and substantially reduce the risk of chronic rejection. This shifted attention towards slowly progressing fibrotic changes that can contribute to the decline of graft function.     

Chronic tissue remodeling is histologically characterised by the appearance of Interstitial Fibrosis and Tubular Atrophy (IFTA) and glomerulosclerosis. In recent years, works studying the mechanisms leading to these pathologies have been carried out \cite{Liu2011,Li2014}. In particular, macrophages have recently been identified as a key player in the inflammation and fibrosis process \cite{Tang2019}. Depending on their phenotype (``M1-like'' or ``M2-like''), macrophages can be pro or anti-inflammatory and they also play a role in the activation of fibroblasts inducing IFTA and glomerulosclerosis.

\begin{figure*}[t!]
	\centering
	\begin{subfigure}[t]{0.3\linewidth}
        \includegraphics[width=\linewidth]{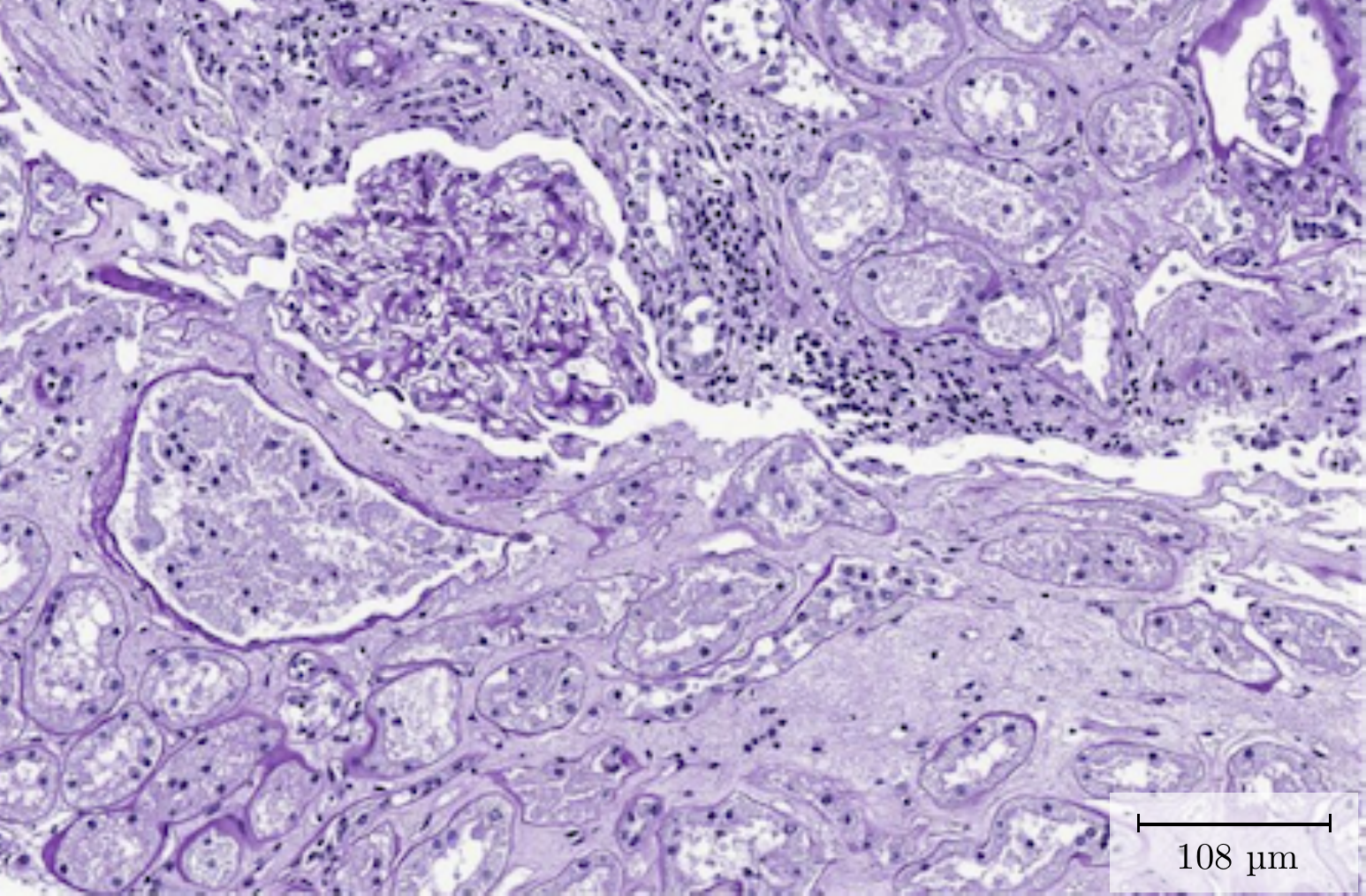}
        \caption{Periodic Acid Schiff (PAS)}
%        \label{}
    \end{subfigure}
	\hspace{0.3cm}
	\begin{subfigure}[t]{0.3\linewidth}
        \includegraphics[width=\linewidth]{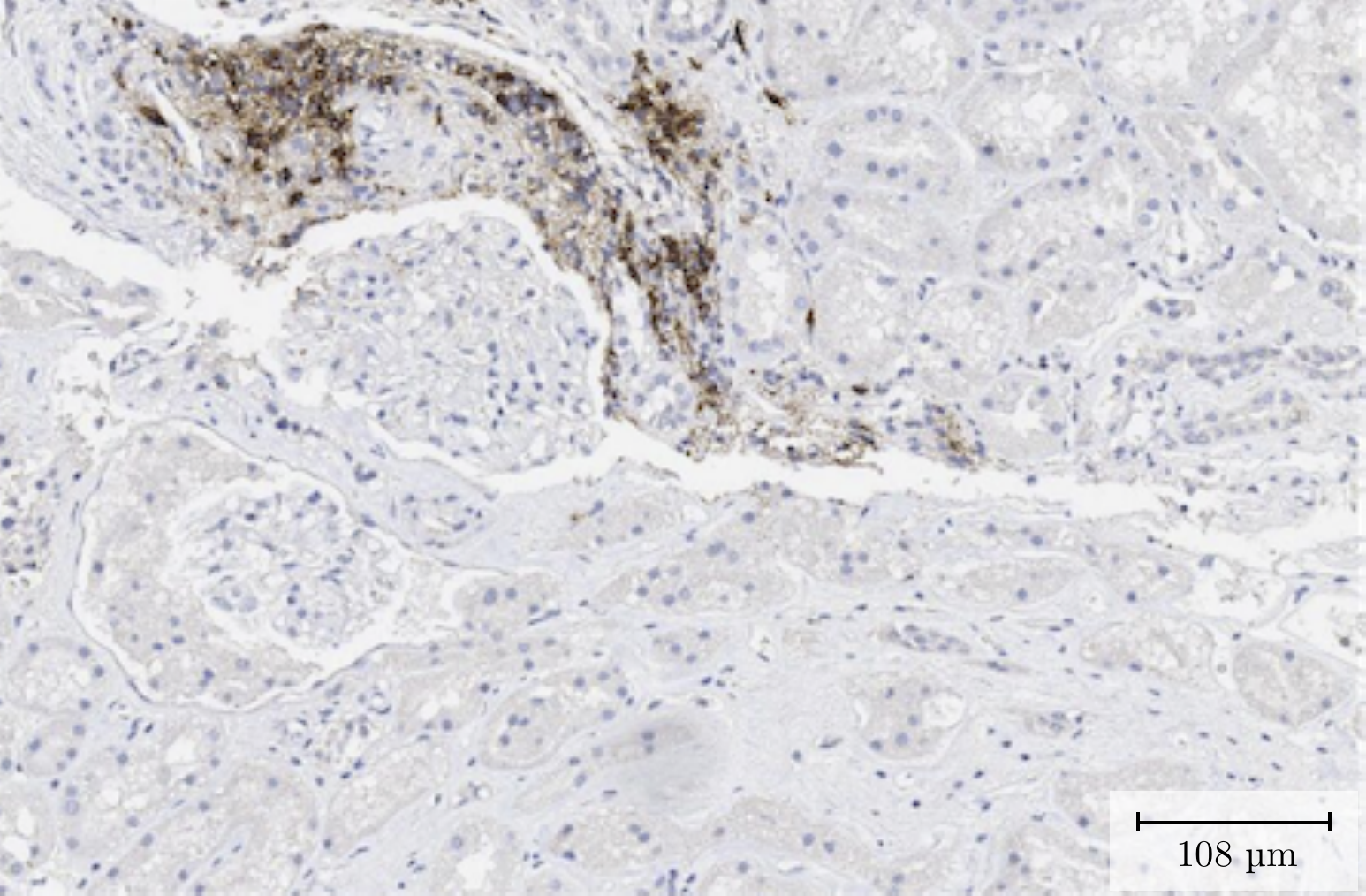}
        \caption{CD3}
%        \label{}
    \end{subfigure}
    \hspace{0.3cm}
	\begin{subfigure}[t]{0.3\linewidth}
        \includegraphics[width=\linewidth]{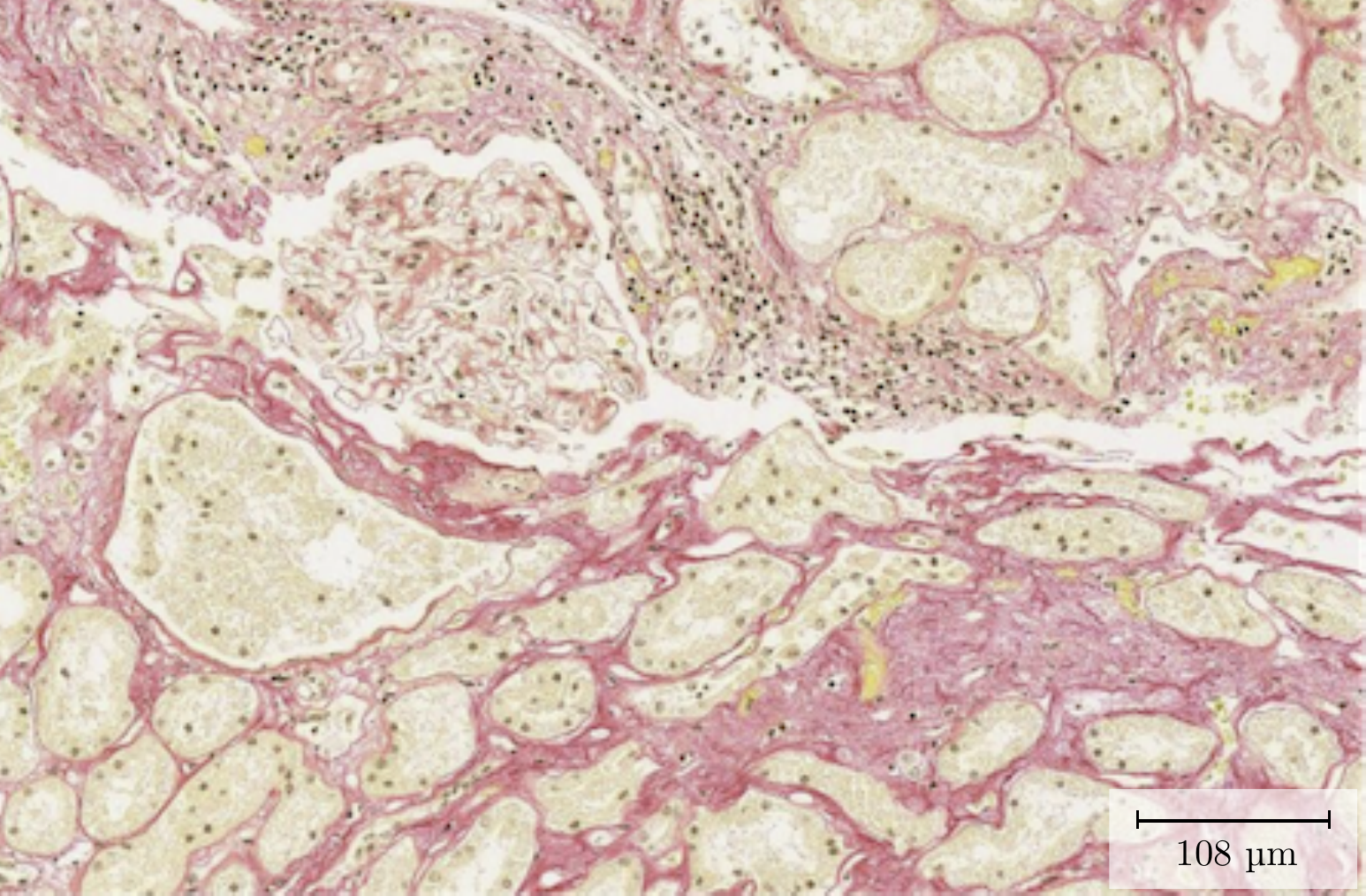}
        \caption{Sirius Red}
%        \label{}
    \end{subfigure}
	\caption{An example of three consecutive WSIs of a kidney nephrectomy sample with three common stains. Each staining provides different information on the tissue: general structural information in PAS, distribution of T lymphocytes in CD3, specific structures such as collagen or muscular fibres in Sirius Red.}
	\label{Fig: example multistained WSI}
\end{figure*}

A common approach in histopathology is the visual evaluation of consecutive, differently stained, biopsy sections by trained pathologists. Each stain provides specific information on the tissue (see Fig.\ \ref{Fig: example multistained WSI}) and the pathologist integrates this information into a written report.

With the emergence of system biomedicine, there has been an increasing trend to study complex mechanisms based on quantitative data such as inflammation \cite{Brasen17, Casper18}, cancer clonal evolution \cite{Angelova18}, or immune reactions \cite{Johnson19}. In this context, Whole Slide Images (WSI) with different stains are studied separately \cite{Farris20,Ginley21}, and the fusion of information from these different stainings is required to obtain a comprehensive data set. Pathologists mentally perform this fusion while analysing a piece of tissue. This trivial task for trained pathologists, commonly referred to as slide registration, is highly complex for computers and requires specifically designed algorithms.

WSI registration algorithms should take into account several specificities intrinsic to histopathology: the tissue shape and orientation of two consecutive slides may vary because of the sample preparation (fixation, embedding, sectioning, etc.); the composition of the tissue between two slides can vary significantly as the cells and structures may appear, disappear, or have different appearance depending on the sectioning level; finally, the set of stainings used may highlight different structures or cells which results in slides that look quite different (see Fig.\ \ref{Fig: example multistained WSI}).

Algorithms used for WSI registration in the literature usually apply non-rigid deformations resulting in visually pleasing registration. Most methods use the mutual information similarity metric to register two WSIs with different stainings \cite{Can08, Mosaliganti06, Mueller11} as it relies on statistical relations between the intensities of two stainings instead of direct correlations. Nevertheless, these methods may fail for stainings with very different appearances as they are only based on raw intensities. To overcome this, Cooper et al.\ \cite{Cooper09} proposed to rely on purely geometric features for the registration. More recently, Song et al.\ \cite{Song14} developed an unsupervised content classification algorithm that computes more complex features describing the structures of each image. Even though these non-rigid methods yield good visual registration, they introduce spatial deformations that change the statistical properties of the neighbouring area, and therefore induce significant bias when extracting geometric features. To avoid this, this article proposes a registration strategy in which structures are matched without tissue deformation, therefore preserving pertinent spatial information between slides. As an alternative to mutual information, Schwier et al.\ \cite{Schwier13} propose to histogram match two WSIs, then threshold them to extract vessels for registration. Gupta et al.\ \cite{Gupta18} use this to calculate a non-rigid registration in order to warp the segmentation mask from one WSI to another. This does not therefore account for glomeruli that appear, disappear, etc between slides and such an approach is unlikely to work as the distance between slides increases (such as with multiple consecutive WSIs). Other works exist that focus on the registration of segmentation masks directly \cite{Kybic14, Kybic15}. These focus on pairs of images, not multiple ($>$ 2) WSIs, nor explicitly result in the fusion of information between slides.  Ultimately, these algorithms also perform non-rigid registration, and therefore fall victim to the same limitation as direct registration of the WSIs. \color{black}

The remainder of this article is organised as follows: 
Section \ref{sec:pipeline} presents the complete analysis pipeline and Section \ref{sec: data} presents the data used for evaluation. A number of experiments are conducted in Section \ref{sec:results} to validate the pipeline: the proposed matching algorithm is validated both independently and in the context of the pipeline; the application of the complete pipeline to four consecutive nephrectomy WSIs; and an illustration of several interesting features that can be computed from such an analysis framework, along with their analysis.

\begin{figure*}[t!]
	\centering
	\includegraphics[width=\textwidth]{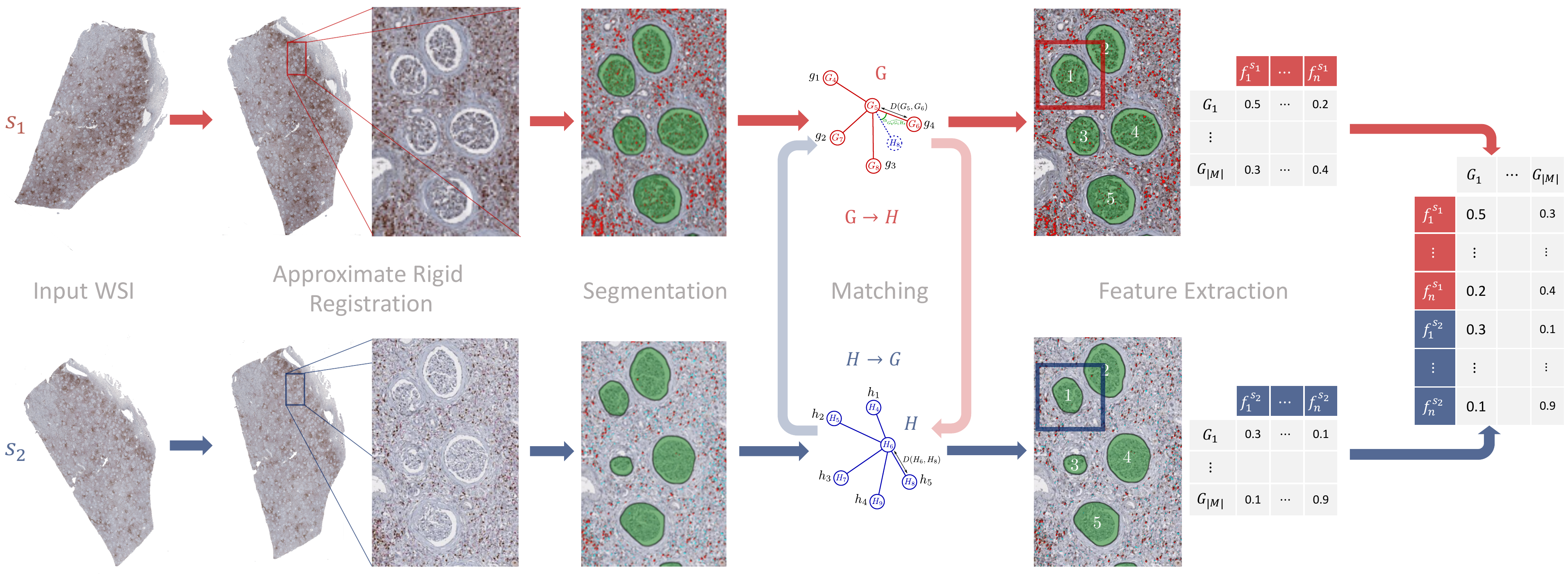}
    \caption{An overview of the proposed analysis framework. First, rigid registration is performed  on two consecutive WSIs with staining $s_1$ and $s_2$ to approximately align the tissue. Glomeruli segmentation, then matching is then performed from $s_1$ to $s_2$ ($G \longrightarrow H$) and from $s_2$ to $s_1$ ($H \longrightarrow G$). Only matchings that are found in both directions are kept. Patches around each matched glomeruli are extracted and for each of them several features are computed. Feature matrices from both stainings are concatenated so that in each column (i.e.\ for each matched glomeruli) we have features extracted from the two stainings $s_1$ (in red) and $s_2$ (in blue).}
    \label{fig:flowchart}
\end{figure*}

%%%%%%%%%%%%%%%%%%%%%%%%%%%%%%%%%%%%%%%%%%%%%%%%%%%%%%%%%%%%%%%%%%%%%%%%%%%%%%%%%%%%%%%%%% Proposed approach %%%%%%%%%%%%%%%%%%%%%%%%%%%%%
%%%%%%%%%%%%%%%%%%%%%%%%%%%%%%%%%%%%%%%%%%%%%%%%%%%%%%%%%%%%%%%%%%%%%

\section{Analysis Framework}
\label{sec:pipeline} 

This article presents an automatic pipeline to analyse histology slides from patients with chronic renal graft rejection. Many features quantifying the inflammation can be extracted from this pipeline and used by pathologists for diagnosis purposes. More complex features, such as spatial correlations between cell populations, can also be extracted to describe the tissue state and help researchers, clinicians and pathologists to better understand the mechanisms leading to IFTA and glomerulosclerosis.

Instead of applying non-rigid registration, 
we propose to merge the information from consecutive slides by finding common landmarks across stainings and locally superimposing the regions from each staining around these landmarks. Glomeruli are spherical structures with a diameter of around $\SI{150}{\micro\metre}$ and are one of the key functional units of the kidney, responsible for the filtration of primary urine from the blood. 
They are thereby good candidates for landmarks as they have a high probability to be present in several consecutive slides (the average slice of tissue is $\SI{3}{\micro\metre}$ thick) and they present an isotropic structure making them easily detected in each slide whatever the cutting direction.

In order to merge the information from several consecutive slides, we propose to match the glomeruli across slices and then locally superimpose each glomerulus neighbourhood to perform the multi-stain analysis. Thus, our framework is four fold: 1) rigid registration to approximately align the tissue, 2) cell and structures segmentation, 3) glomeruli matching, 4) feature extraction from different stainings in the same spatial reference. An overview of this framework is presented in Fig.\ \ref{fig:flowchart}.

Rigid registration and segmentation are open problems in histopathological imaging, and numerous approaches exist. The purpose of this article is not to propose new approaches in these areas, therefore the following assumes that the tissue can be approximately rigidly registered using one of the available algorithms \cite{Lowekamp13, Avants14, Klein10}, and that we have accurate (but not necessarily perfect) segmentation of the cells and glomeruli. In this work a U-Net was used for glomeruli segmentation \cite{Ronneberger15, Lampert19} and stain deconvolution for cell segmentation \cite{Ruifrok01}. More details of these two steps are provided in Section \ref{sec:gcseg} of the Supplementary Material. For alternatives, we refer the reader to recent reviews \cite{GurcanRBE09, Xing16}.

The key contributions of this article are as follows.
\begin{itemize}[noitemsep,topsep=0pt]
    \item A new stain registration strategy to avoid tissue deformation based on glomeruli matching. This matching algorithm is validated on real data showing robust performance (see Section \ref{sec: registration}).
    
    \item An automatic pipeline able to extract quantitative features from consecutive WSIs with different stainings. Combined with the matching algorithm, this pipeline allows for the registration of features from different stainings in the same spatial context, leading to a global multi-stain analysis pipeline.
\end{itemize}

\subsection{Glomeruli Matching}
\label{sec: registration}

%%%%%%%%%%%%%%%%%%%%%%%%%%%%%%%%%%%%%%%%%%%%%%%%%%%%%%%%%%%%%%%%%%
%%%%%%%%%%%%%%%%%%%%% Glomeruli Matching %%%%%%%%%%%%%%%%%%%%%%%%%
%%%%%%%%%%%%%%%%%%%%%%%%%%%%%%%%%%%%%%%%%%%%%%%%%%%%%%%%%%%%%%%%%%

This section presents a novel glomeruli matching algorithm in order to locally superimpose glomeruli neighbourhoods between slices.

Let $G$ be the set of glomeruli in a WSI and $H$ be the set of glomeruli in a WSI consecutive to it. The cardinality of a set $A$ is denoted $|A|$, such that $|G|$ and $|H|$ are the number of glomeruli respectively in $G$ and $H$. 

Matching $G$ to $H$ can be seen as an inexact graph matching problem. Let $\mathcal{G}=(G, E_G)$ and $\mathcal{H}=(H, E_H)$ be two graphs where $E_G$ (resp.\ $E_H$) is a set of edges between the glomeruli $G$ (resp.\ $H$). The inexact matching problem is defined as
\begin{equation}
\hat{x} = \argmin_{x \in X} \sum_{k} E(G_k,H_{x_k}),
\label{Eq:inexact graph matching}
\end{equation}
where $X\in \mathds{N}^{|G|}$ is the set of all possible matchings from $\mathcal{G}$ to $\mathcal{H}$ and $E:G\times H \rightarrow \mathds{R}$ is a matching energy function.

Inexact graph matching is an NP-complete problem that is usually solved by finding an approximate solution using heuristic search strategies. In this work, the complexity of the global inexact graph matching problem is reduced by incorporating prior knowledge regarding the solution and adopting a subgraph assignment splitting strategy inspired by the work of Raveaux et al.\ \cite{RaveauxPRL10}.

The largest contribution to the complexity of general graph matching comes from the combinatorial problem of trying to match every vertex in $G$ to every vertex in $H$ regardless of their respective spatial location.
Herein, it is assumed that the position of the same glomerulus in two consecutive slides should be similar relative to the surrounding tissue, since in the previous step the tissues in both slides have been approximately rigidly registered.

Based on this observation, the global inexact graph matching problem of $\mathcal{G}$ to $\mathcal{H}$ is transformed into $|G|$ sub-graph ($\text{sub}$) assignment problems. Each sub-problem, i.e.\ for each $g \in G$, attempts to find the best matching subset of vertices in $H$, the number of candidate vertices of $H$ is reduced to those having similar spatial positions to $g$ relative to the tissue.

In the following, the general matching strategy is first developed, then the assignment energy used to match two glomeruli is presented.

\subsubsection{From Global to Local Matching}

Let $G$ and $H$ be embedded in $\mathds{R}^2$. We define the set of edges of both graphs such that $E_G = \{(x,y) \in G^2, D(x,y) \leq d_\text{sub}\}$ and $E_H = \{(x,y) \in H^2, D(x,y) \leq d_\text{sub}\}$, where $D:\mathrm{R}\times \mathrm{R} \rightarrow \mathrm{R}$ is a function returning the Euclidean distance between two points (vertices) and $d_\text{sub}\in \mathds{R}$ is the maximum length of an edge in the subgraph.

\begin{figure*}[t]
	\centering
	\begin{subfigure}[c]{0.32\linewidth}
        \includegraphics[width=\textwidth,frame]{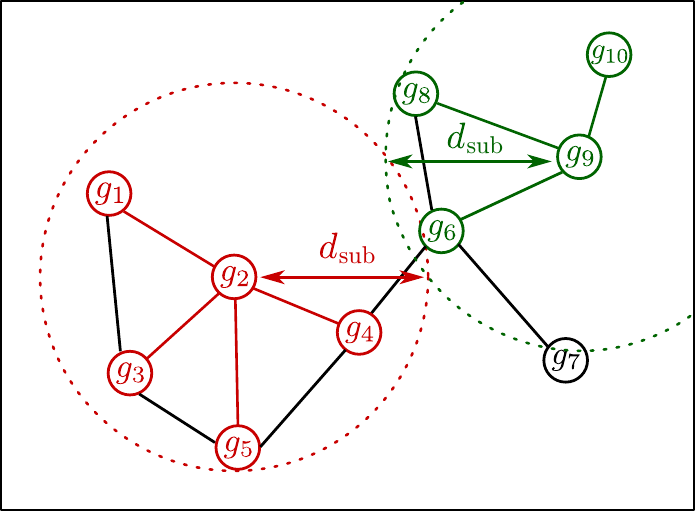}
        \caption{}
        \label{subfig:matchinga}
    \end{subfigure}
	\begin{subfigure}[c]{0.32\linewidth}
        \includegraphics[width=\textwidth,frame]{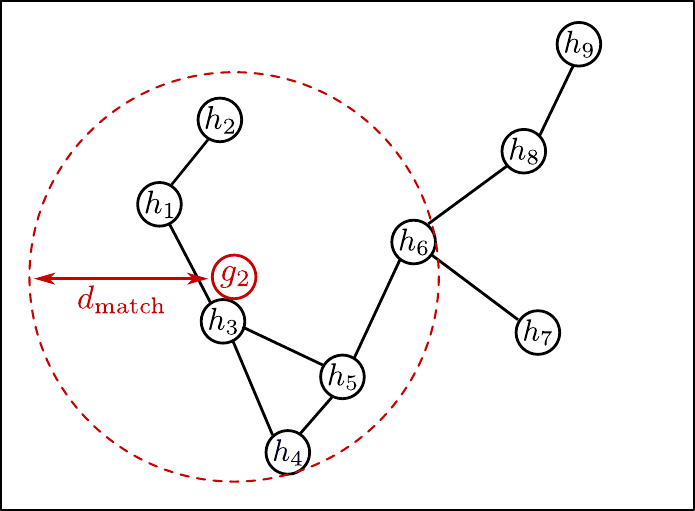}
        \caption{}
    \end{subfigure}
    \begin{subfigure}[c]{0.32\linewidth}
        \includegraphics[width=\textwidth,frame]{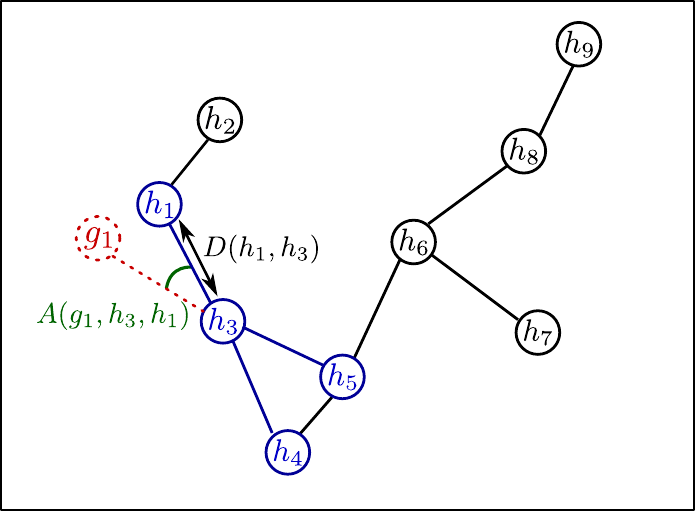}
        \caption{}
    \end{subfigure}
    
    \caption{An illustration of the glomeruli matching steps of two graphs $\mathcal{G}$ (a) and $\mathcal{H}$ (b). (a) Example of two subgraphs $\mathcal{G}^\text{sub}_2$ in red and $\mathcal{G}^\text{sub}_9$ in green of $\mathcal{G}$. (b) The set of vertices of $H$ that can be matched to $g_2$: $\mathcal{N}_H^{d_\text{match}}(g_2)= \{h_1, h_2, h_3, h_4, h_5, h_6\}$. (c) The energy $E_{\text{nb}}^{g_2,h_3}(g_1, h_1)$ is computed based on the angle $A(g_1,h_3,h_1)$ and the distances $D(h_1,h_3)$ and $D(g_1,g_2)$.
    In the example of matching $\mathcal{G}^\text{sub}_2$ to $\mathcal{H}^\text{sub}_3$, the matching energy for $N=3$ would be $E_{\text{match}}(g_2, h_3) = E_\text{nb}^{g_2h_3}(g_1,h_1) + E_\text{nb}^{g_2h_3}(g_5,h_4) + E_\text{nb}^{g_2 h_3}(g_4,h_5)$.
    }
    \label{fig:glomeruli matching steps}
\end{figure*}

Instead of finding a global matching, i.e.\ Equation \eqref{Eq:inexact graph matching}, that could lead to the matching of glomeruli far from each other in consecutive WSIs, the problem is reduced to $|G|$ sub-problems defined by finding for each vertex $g\in G$, its best match $\hat{h} \in H$, among all vertices of $H$ that are close to $g$, such that
\begin{equation}
\label{Eq:sub matching problem}
\hat{h}= \argmin_{h \in \mathcal{N}_H^{d_{\text{match}}}(g)} E_\text{match}(g,h),
\end{equation}
where $\mathcal{N}_H^{d_{\text{match}}}(g) = \{ h\in H, D(h,g) \leq d_{\text{match}}\}$ is the set of vertices of $H$ that have spatial positions similar to $g$ (in the rigidly registered image) and $ E_\text{match}(g,h)$ is the energy of matching $h$ to $g$ that will be defined in the next section. An illustration of $\mathcal{N}_H^{d_{\text{match}}}(g) $ is presented in Fig.\ \ref{fig:glomeruli matching steps}.

Since a glomerulus of $H$ can only be matched to one glomerulus of $G$, the matching with the lowest energy $E_\text{match}$ is retained for each $g$.

\begin{figure*}[t]
	\centering
	\begin{subfigure}[c]{0.3\linewidth}
        \includegraphics[width=\textwidth,frame]{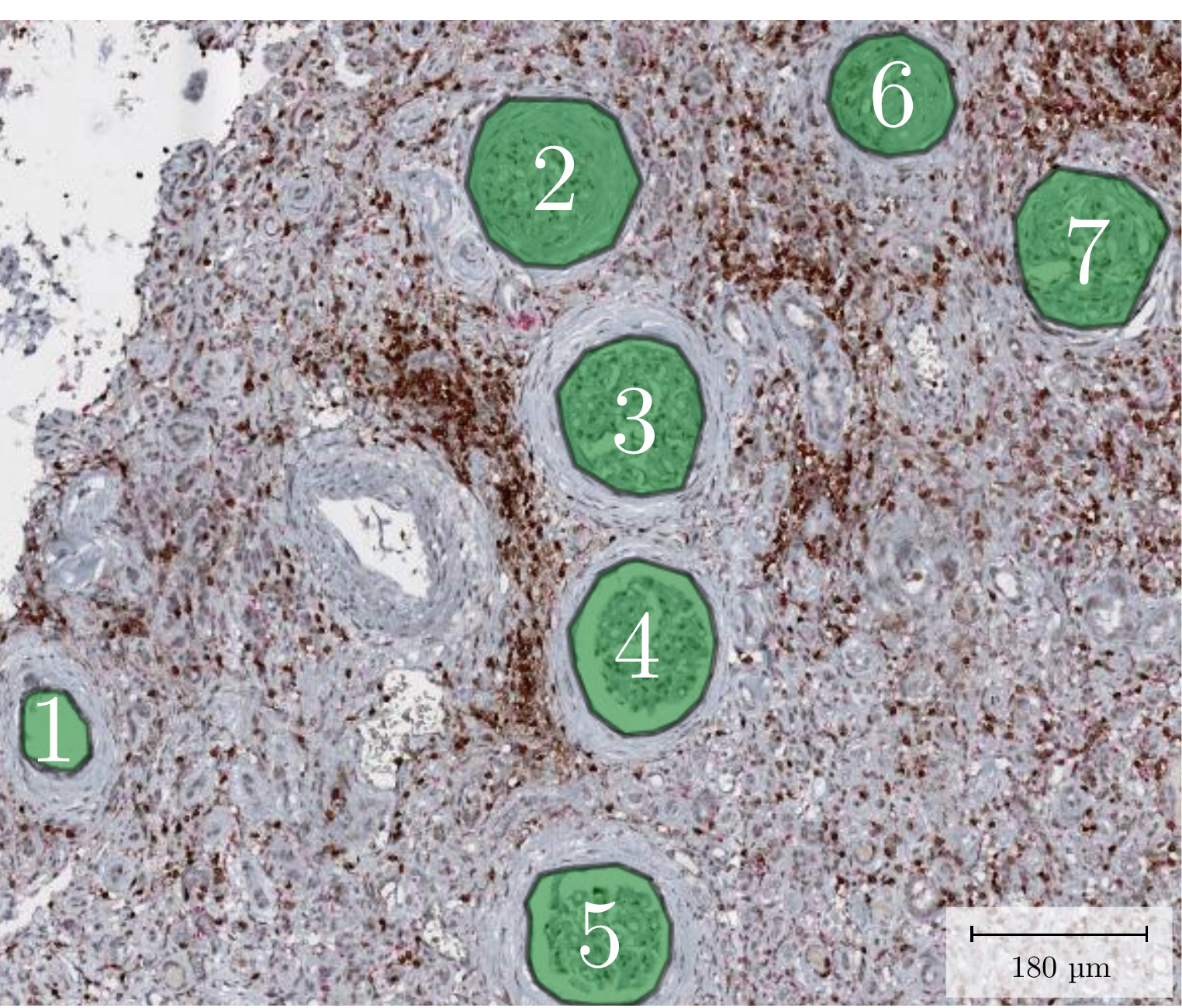}
        \caption{}
    \end{subfigure}
	\hspace{0.2cm}
	\begin{subfigure}[c]{0.3\linewidth}
        \includegraphics[width=\textwidth,frame]{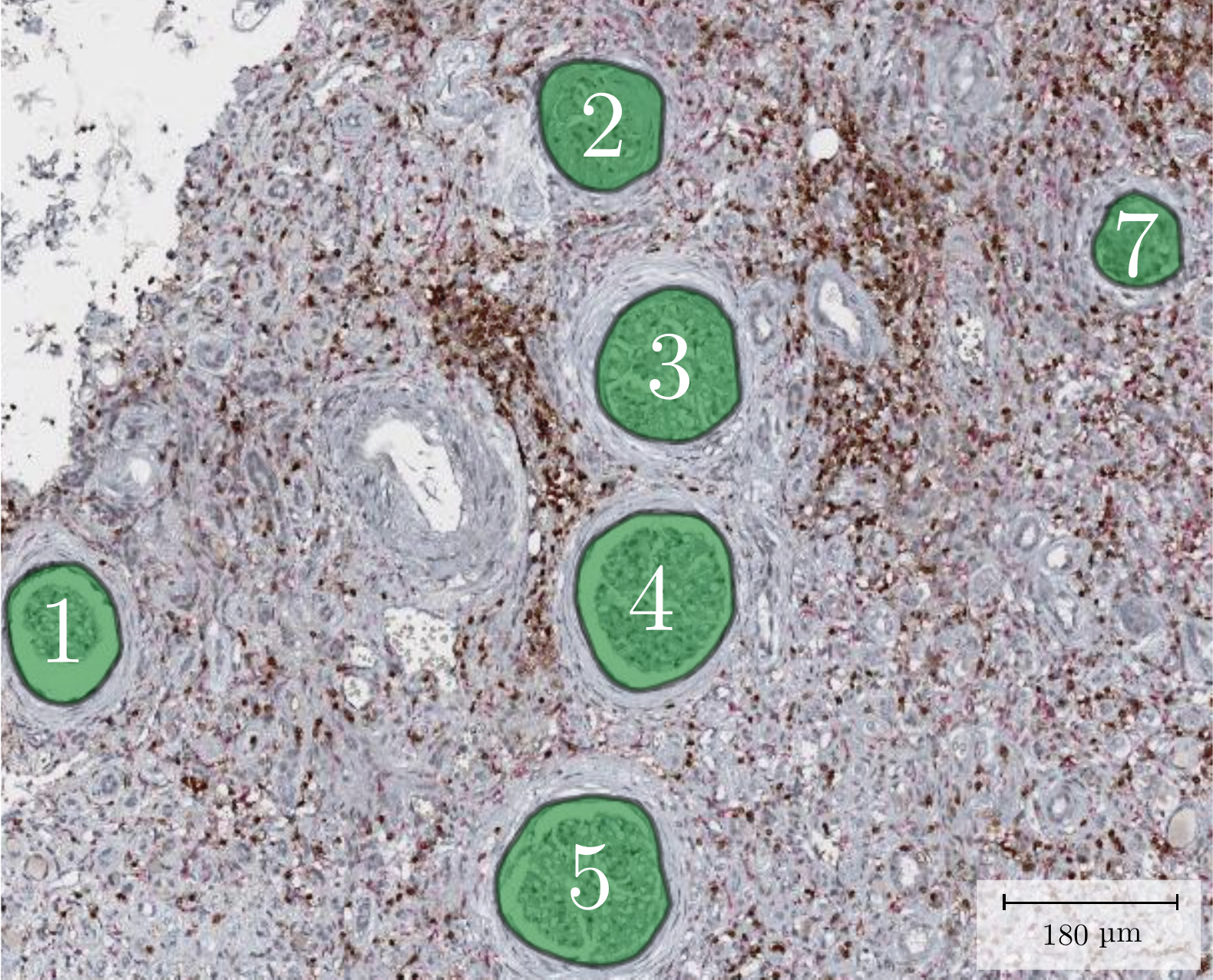}
        \caption{}
        \label{subfig:matchingb}
    \end{subfigure}
    \caption{Illustrating glomeruli matching on two consecutive WSIs. Consecutive glomerulus cuts may present different shapes and sizes (cf.\ glomeruli number 1 and 7) but their position in the tissue relative to other glomeruli is similar.}
    \label{fig:glomeruli_matching_illustration}
\end{figure*}

\subsubsection{Assignment of Glomeruli Neighbourhood}

It can be observed in WSIs that although the shape and size of a glomerulus slice may vary significantly between consecutive slides, its position relative to neighbouring glomeruli is relatively constant (see Fig.\ \ref{fig:glomeruli_matching_illustration}).
To constrain the matching strategy with this observation, the matching energy of two glomeruli slices $g$ and $h$ is defined to be the minimal assignment energy of their respective neighbourhoods. More formally, let $\mathcal{G}^\text{sub}_i = (G_i^\text{sub}, E_{G_i}^\text{sub})$ be a subgraph centred on $g_i \in G$ such that
\begin{align*}
G_i^\text{sub} &= \{g_k\in G, (g_i, g_k) \in E_G\} \cup  g_i,\\
E_{G_i}^\text{sub} &= \{ (x, g_i) \in E_G, \ x \in G\}.
\end{align*}
Fig.\ \ref{fig:glomeruli matching steps} presents examples of such sub-graphs.

Let $\tilde{G}_i^\text{sub} = G_i^\text{sub} \setminus g_i$ be the set of vertices connected to $g_i$. The assignment of $\tilde{G}_i^\text{sub}$ to $\tilde{H}_j^\text{sub}$ is defined such that
\begin{equation}
E_{\text{match}}(g_i, h_j) = \min_{f\in F^{ij}} \sum_{k=1}^N RF_{k} \left(E_{\text{nb}}^{g_ih_j}(g, f(g)),\ g \in \tilde{G}_i^\text{sub}\right),
\end{equation}
where $F^{ij}$ is the set of all possible mappings of a vertex in $\tilde{G}_i^\text{sub}$ to a vertex of $\tilde{H}_j^\text{sub}$, and $RF_{k}$ is the rank filter of order $k$ such that $RF_{1}\left((a_i)_{i\in [1,N]}\right)$ is the minimum and $RF_{N}\left((a_i)_{i\in [1,N]}\right)$ is the maximum. The term $E_{\text{nb}}^{g_ih_j}$ is the energy of assigning neighbouring ($\text{nb}$) vertices such that
\begin{equation}
E_{\text{nb}}^{g_ih_j}(g,h) = \frac{1}{90} A(g,h_j,h) + \frac{|D(g_i,g) - D(h_j,h)|}{D(g_i,g)},
\end{equation}
where $A(g, h_j, h) \in [0, 180]$ is the angle $\widehat{gh_jh}$ in degrees (see Fig.\ \ref{fig:glomeruli matching steps}).

As such, the energy $E_{\text{match}}(g_i, h_j)$ is the sum of the $N$ neighbour associations with the lowest $E_{\text{nb}}$ with $N \leq |\tilde{G}_i^\text{sub}|$. The parameter $N$ allows flexibility in the neighbourhood pattern matching, which is necessary as some neighbours can appear or disappear between two slides. As $N$ increases, the matching is less flexible.
This matching strategy is performed bidirectionally, i.e.\ from $G$ to $H$ and $H$ to $G$, to increase its robustness. The matchings that are consistent between the two are kept to form the set of matched glomeruli $M$.

\subsubsection{Parameter values}

The proposed algorithm has two parameters: the maximum distance within which a match can be found, $d_{\text{match}}$, and the number of neighbour associations to compute the assignment energy, $N$; and a hidden parameter $d_{\text{sub}}$, which is the distance defining the subgraphs.

In practice, $d_{\text{sub}}$ is defined based on the glomeruli distribution and the number of associations $N$ required to compute the assignment energy. As the assignment energy is defined based on $N$ associations, most glomeruli should have at least $N$ neighbours. In practice $d_{\text{sub}}$ is defined such that most of the glomeruli in the image have at least $N+1$ neighbours to take into account the appearance and disappearance of glomeruli between consecutive slides. The values of N and $d_\text{match}$ should be set experimentally depending on the density of glomeruli in the WSI.

The robustness of this algorithm to centroid shifts and disappearance (which can be caused by the natural dissection of a glomeruli, as well as segmentation errors) was experimentally assessed on synthetic data (see the accompanying supplementary material).

\subsection{Feature Extraction}
\label{sec: feature extraction}

\begin{figure*}[t]
	\centering
	\begin{subfigure}[t]{0.2\linewidth}
        \includegraphics[width=\linewidth]{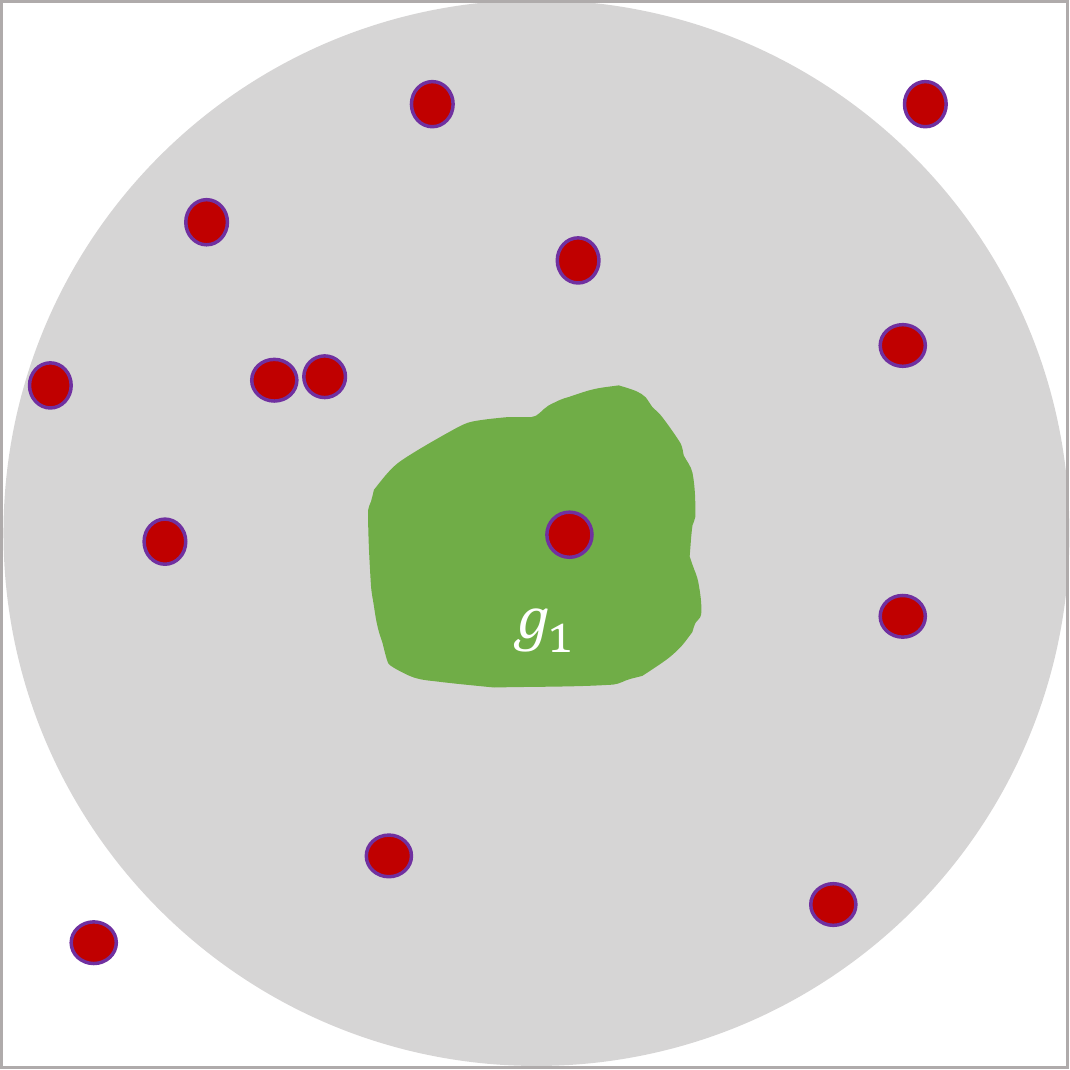}
        \caption{Stain $G$}
%        \label{}
    \end{subfigure}
	\hspace{0.05cm}
	\begin{subfigure}[t]{0.2\linewidth}
        \includegraphics[width=\linewidth]{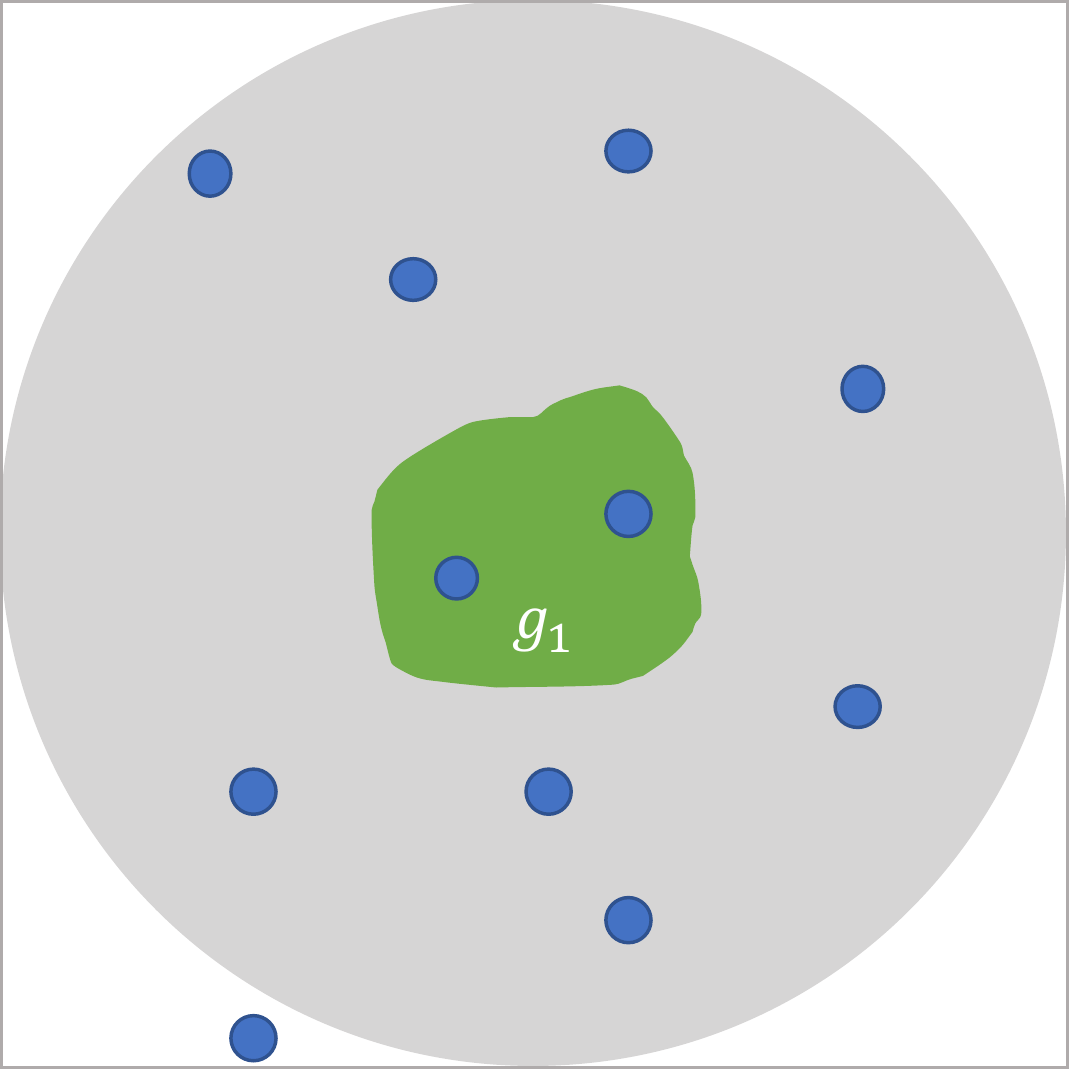}
        \caption{Stain $H$}
%        \label{}
    \end{subfigure}
    \subfloat{
        \raisebox{7ex}{$\Longrightarrow$}
        }
    %\end{subfigure}
    \addtocounter{subfigure}{-1}
    \begin{subfigure}[t]{0.27\linewidth}
    \centering
        \raisebox{2ex}{\includegraphics[width=\linewidth]{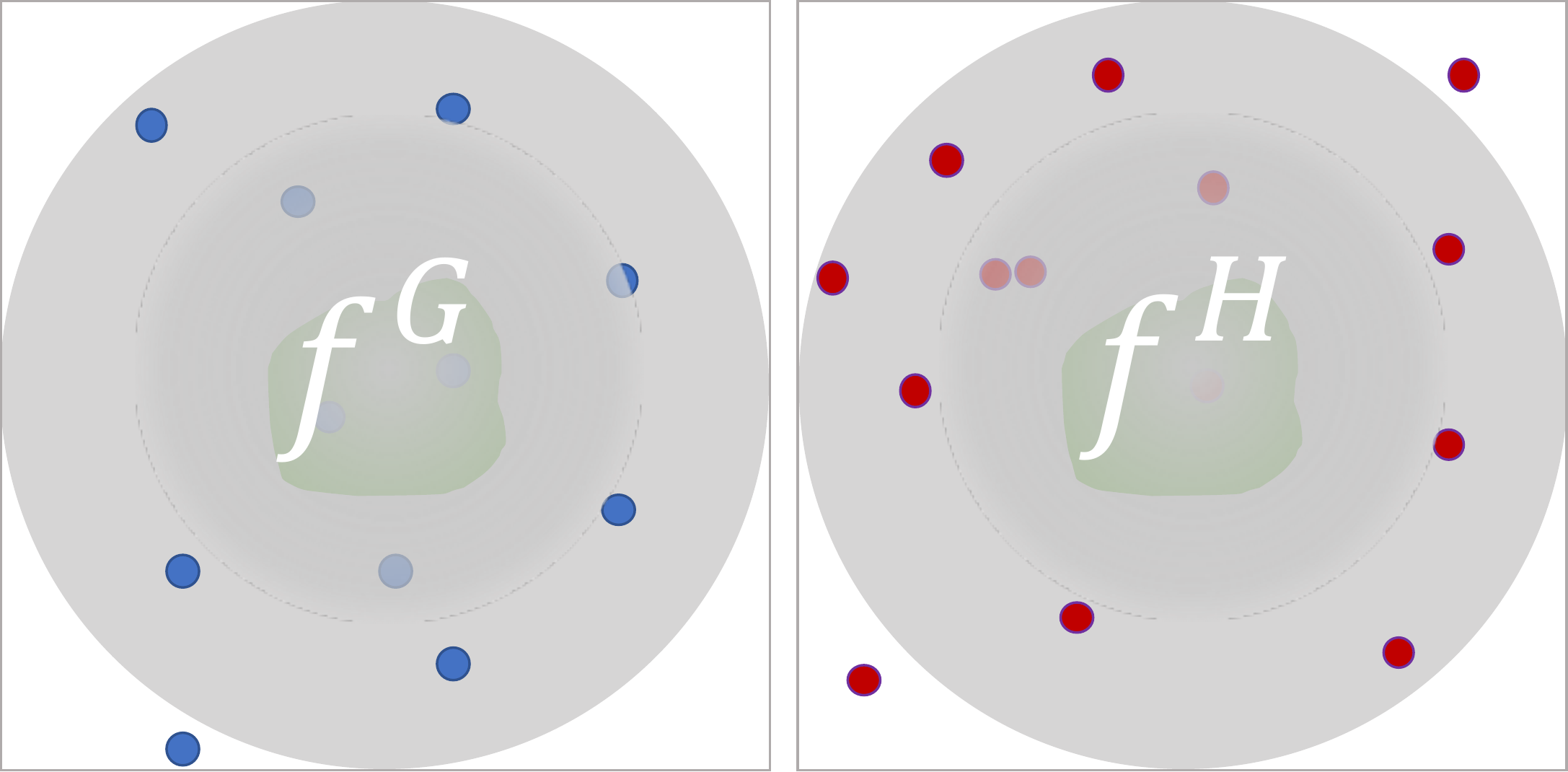}}
        $\mathbf{F} = [f^{G}, f^{H}]$
        \caption{Multi-WSI features}
%        \label{}
    \end{subfigure}
    \hspace{0.05cm}
    \begin{subfigure}[t]{0.2\linewidth}
    \centering
        \includegraphics[width=\linewidth]{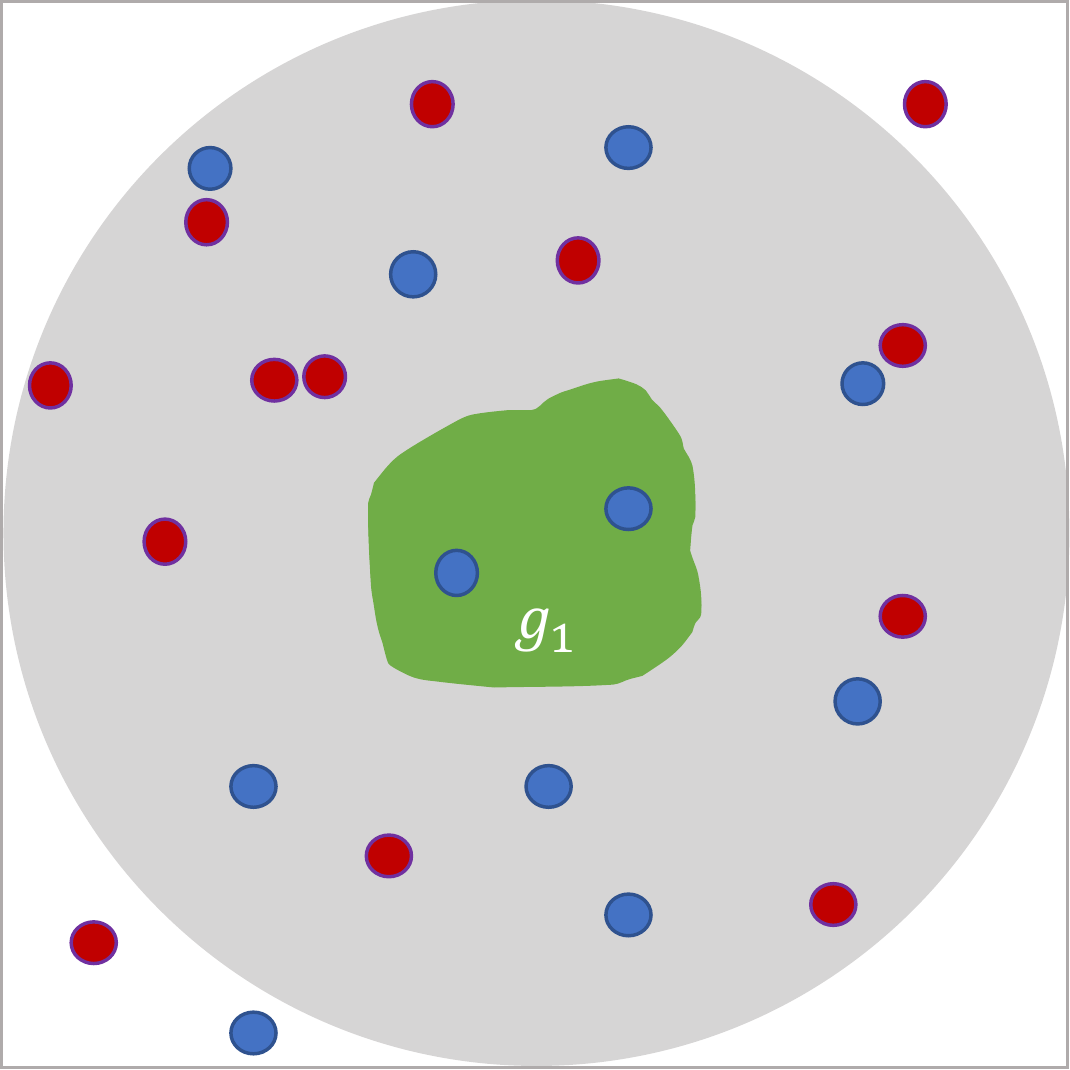}
        $f^{GH}$
        \caption{Intra-WSI features}
%        \label{}
    \end{subfigure}
	\caption{After matching, the same glomerulus (in green) and surrounding tissue in two different stain modalities (a) and (b), the cells of different types are segmented (in red and blue). Two types of features can therefore be computed: features depending on each single WSI concatenated to obtain features relevant to the same glomerulus (c); and features that combine information from both stain modalities (d).}
	\label{fig:twofeaturetypes}
\end{figure*}

Once segmentation and matching across WSIs is complete, the following two types of features, that integrate information derived from different stains in the same glomerulus neighbourhood, can be extracted from the corresponding segmentations, as illustrated in Fig.\ \ref{fig:twofeaturetypes}.
\begin{description}[leftmargin=*]
    \item[Multi-WSI] features derived from multiple single WSIs, for example mean M0 macrophage (CD68) or M2 macrophage (CD163) densities inside each glomerulus.
    \item[Intra-WSI] features, that combine information derived from multiple WSIs, for example, the mean distance from M0 macrophages (CD68) to a subtype of M2 macrophages (CD163).
\end{description}

\section{Dataset}\label{sec: data}

Tissue samples were collected from four patients who underwent allograft nephrectomy for various reasons,  described in Section \ref{sec:pc} of the Supplementary Material.    
Each paraffin-embedded sample was cut into four consecutive \SI{3}{\micro\metre} thick sections, each being stained with one of the following combination of immunohistochemistry markers using an automated staining instrument (Ventana Benchmark Ultra) (the details of the staining material are given in Section \ref{sec:sc} of the Supplementary Material): CD3-CD68 (T cells \& macrophage lineage marker), CD3-CD163 (T cells \& M2-like macrophages), CD3-CD206 (T cells \& M2-like macrophages), or CD3-MS4A4A (T cells \& M2-like macrophages). The $4$ M2-like macrophage stainings detect different subsets of M2 macrophages polarised along the large spectrum of alternatively activated (``M2-like'') macrophages \footnote{This dataset was built in the context of a research project focusing on the role of macrophages in IFTA and glomerulosclerosis, hence the non-routine macrophage-centred stains. The proposed pipeline allows for the comparison of the spatial distributions of macrophages, which is of great interest for this specific project. Nonetheless, this pipeline is generic and any stain may be used.}.

\begin{figure*}[t]
	\centering
	\begin{subfigure}[t]{0.235\linewidth}
        \adjustbox{valign=T}{\includegraphics[width=\linewidth,frame]{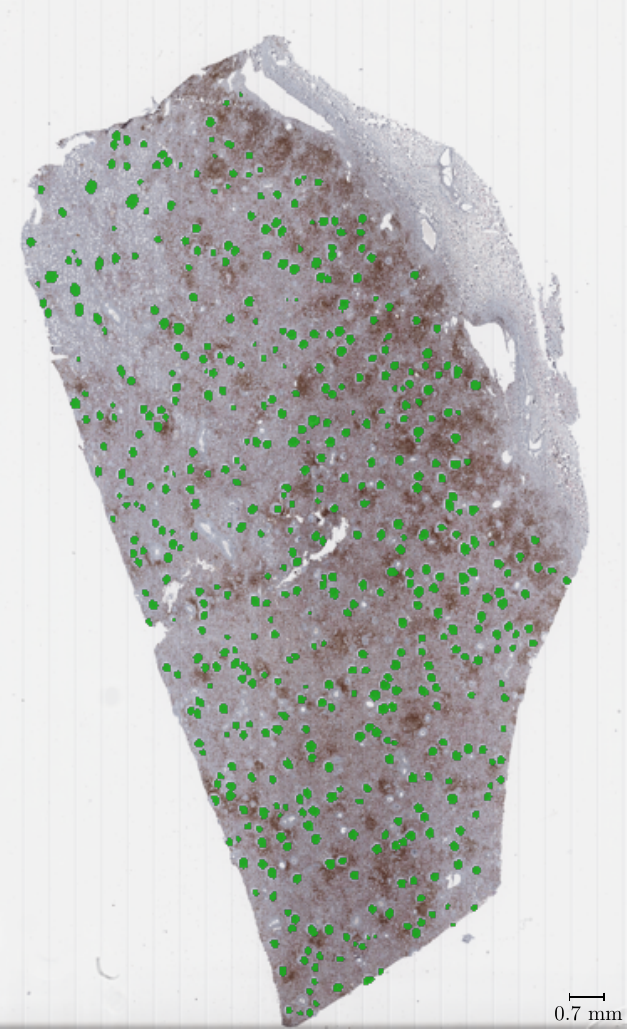}}
        \caption{CD3-CD68}
%        \label{}
    \end{subfigure}
	\hspace{0.01cm}
	\begin{subfigure}[t]{0.22\linewidth}
        \adjustbox{valign=T}{\includegraphics[width=\linewidth,frame]{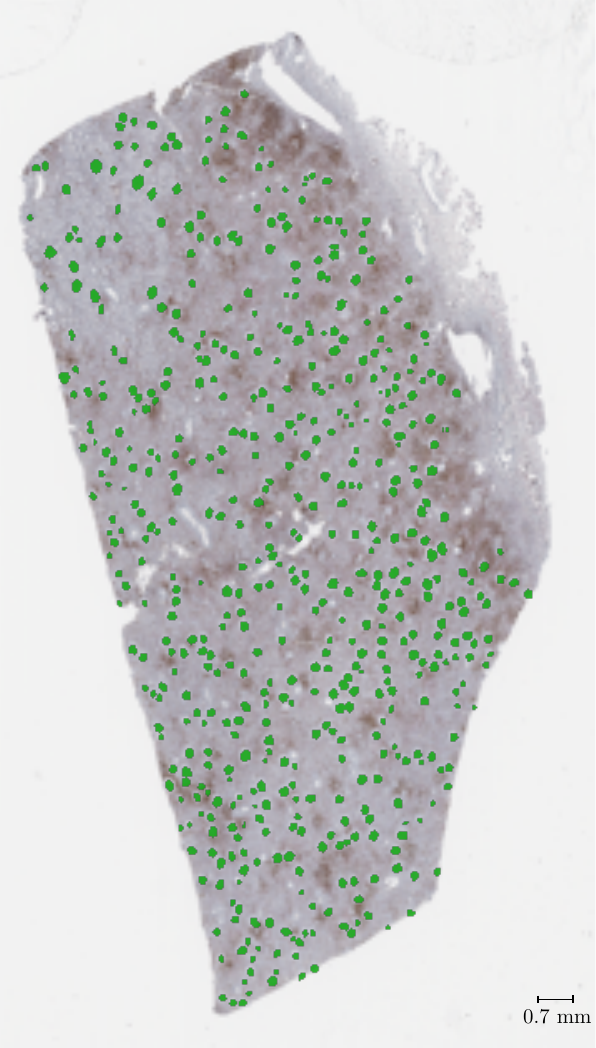}}
        \caption{CD3-CD163}
%        \label{}
    \end{subfigure}
    \hspace{0.01cm}
	\begin{subfigure}[t]{0.23\linewidth}
        \adjustbox{valign=T}{\includegraphics[width=\linewidth,frame]{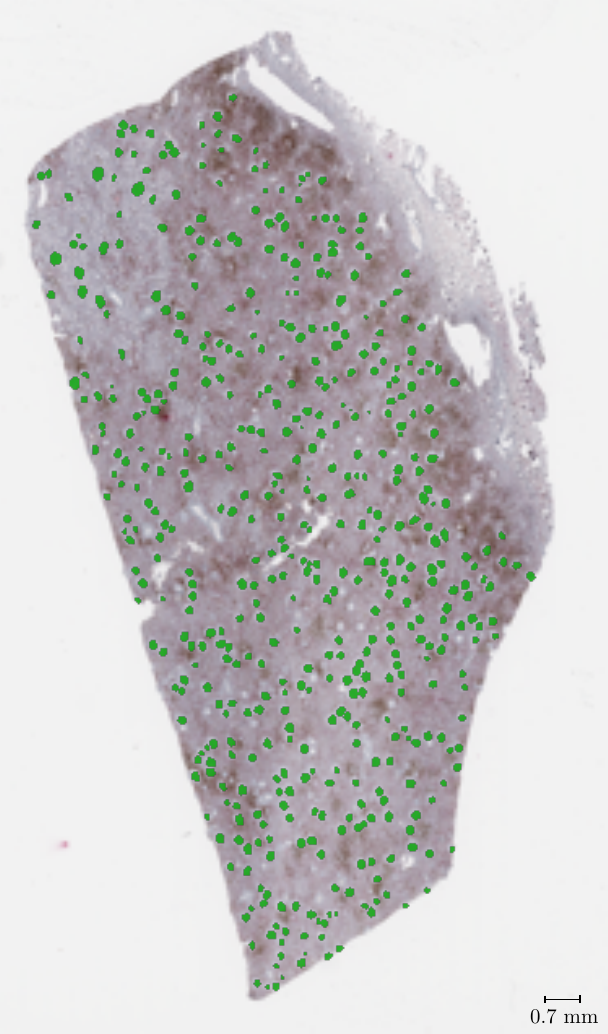}}
        \caption{CD3-CD206}
%        \label{}
    \end{subfigure}
    \hspace{0.01cm}
	\begin{subfigure}[t]{0.23\linewidth}
        \adjustbox{valign=T}{\includegraphics[width=\linewidth,frame]{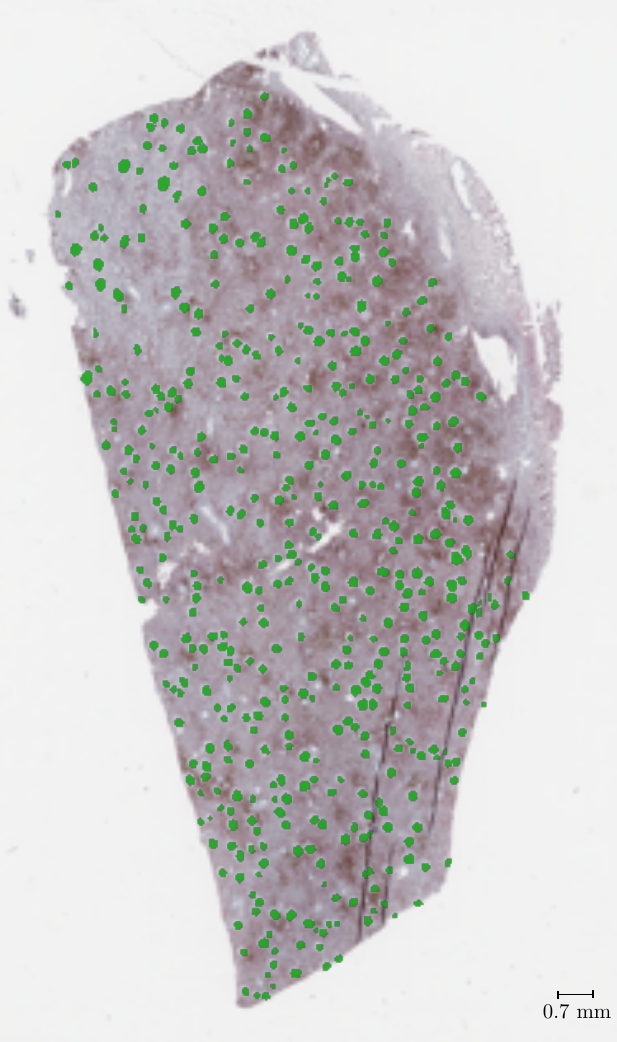}}
        \caption{CD3-MS4A4A}
%        \label{}
    \end{subfigure}
	\caption{Fully annotated consecutive kidney nephrectomy WSIs used in this study (patient 1, see Table \ref{tbl:glomerulinumbers}). Each green disk is an individual glomerulus.}
	\label{Fig: example multistained WSI study}
\end{figure*}

\begin{figure*}[h!]
	\centering
	\begin{subfigure}[t]{0.16\linewidth}
        \includegraphics[width=\linewidth,frame]{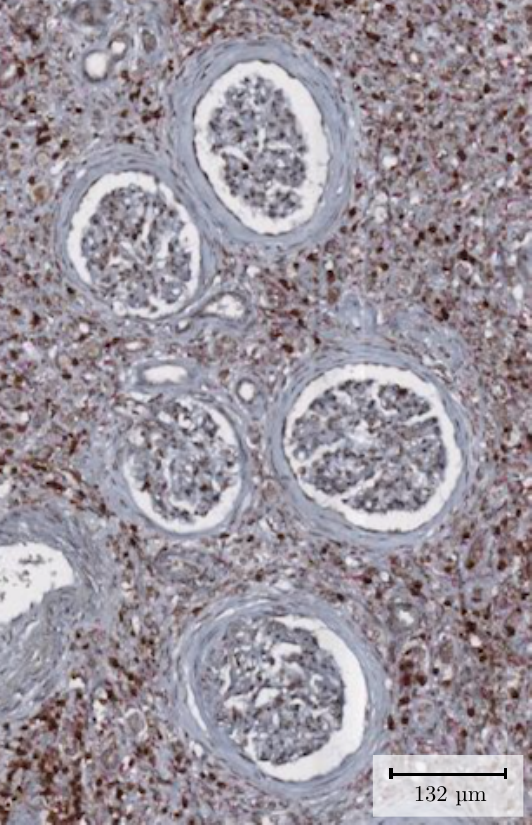}
%        \label{}
    \end{subfigure}
	\hspace{0.05cm}
	\begin{subfigure}[t]{0.16\linewidth}
        \includegraphics[width=\linewidth,frame]{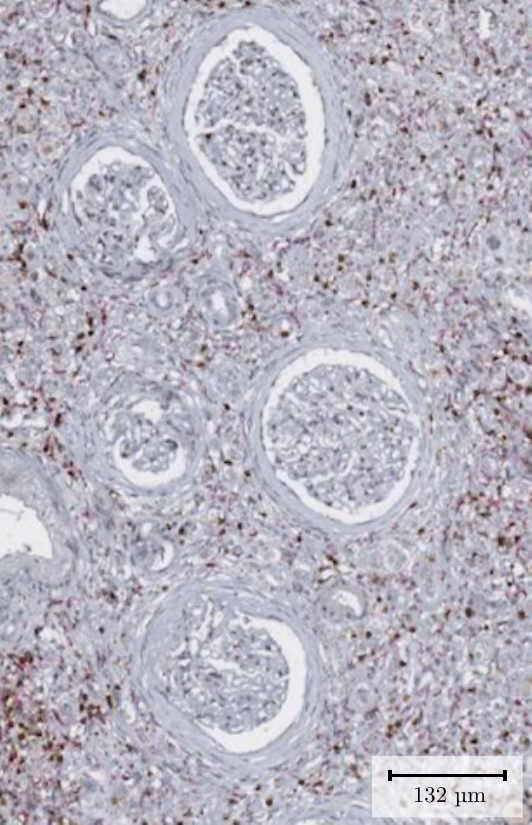}
%        \label{}
    \end{subfigure}
    \hspace{0.05cm}
	\begin{subfigure}[t]{0.16\linewidth}
        \includegraphics[width=\linewidth,frame]{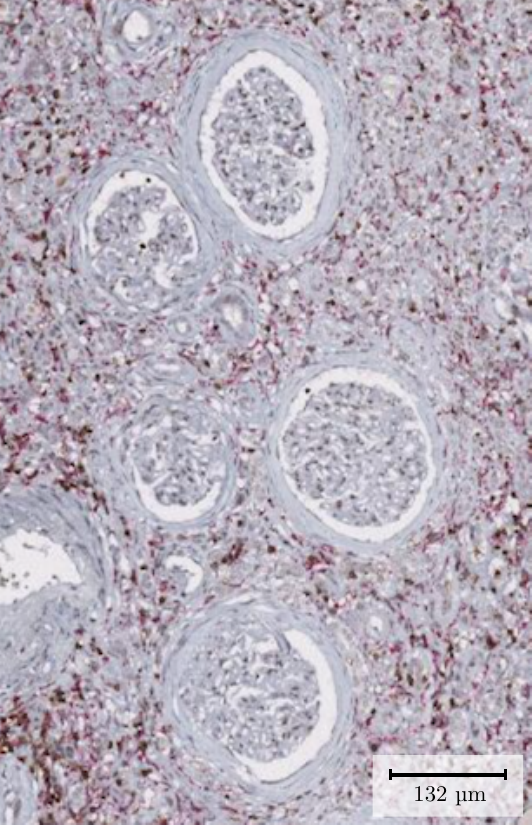}
%        \label{}
    \end{subfigure}
    \hspace{0.05cm}
	\begin{subfigure}[t]{0.16\linewidth}
        \includegraphics[width=\linewidth,frame]{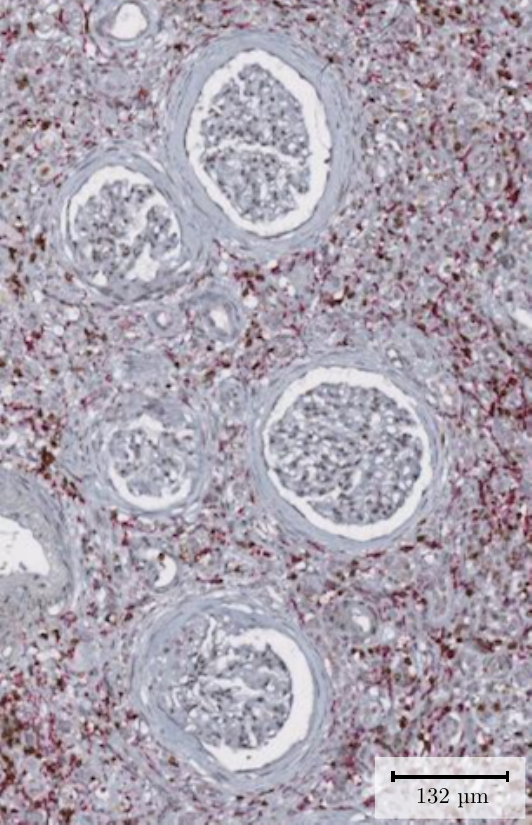}
%        \label{}
    \end{subfigure}
    \\
    \begin{subfigure}[t]{0.16\linewidth}
        \includegraphics[width=\linewidth,frame]{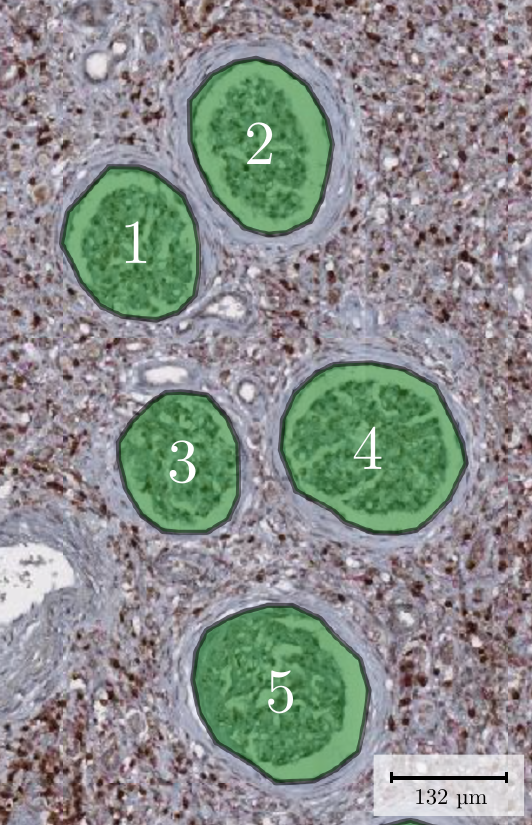}
%        \label{}
    \end{subfigure}
	\hspace{0.05cm}
	\begin{subfigure}[t]{0.16\linewidth}
        \includegraphics[width=\linewidth,frame]{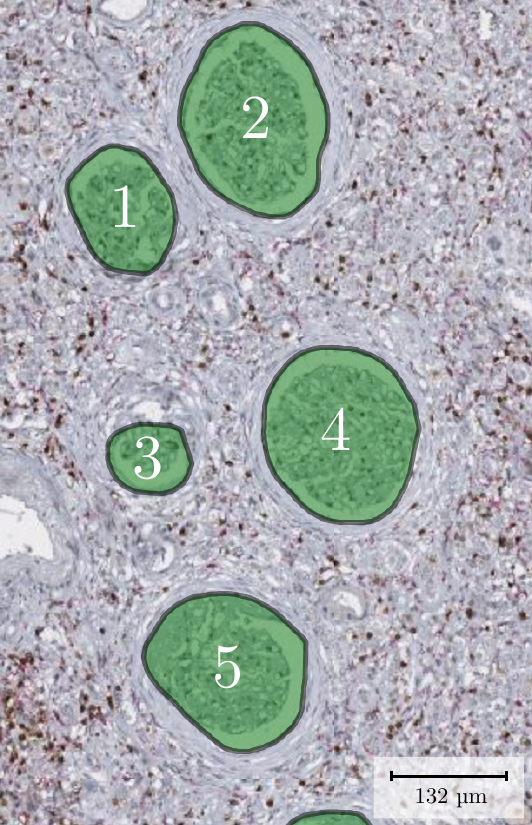}
%        \label{}
    \end{subfigure}
    \hspace{0.05cm}
	\begin{subfigure}[t]{0.16\linewidth}
        \includegraphics[width=\linewidth,frame]{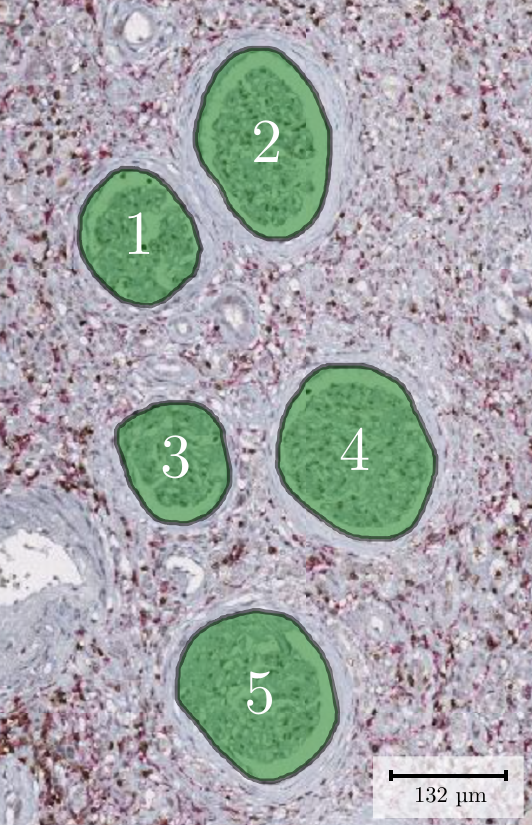}
%        \label{}
    \end{subfigure}
    \hspace{0.05cm}
	\begin{subfigure}[t]{0.16\linewidth}
        \includegraphics[width=\linewidth,frame]{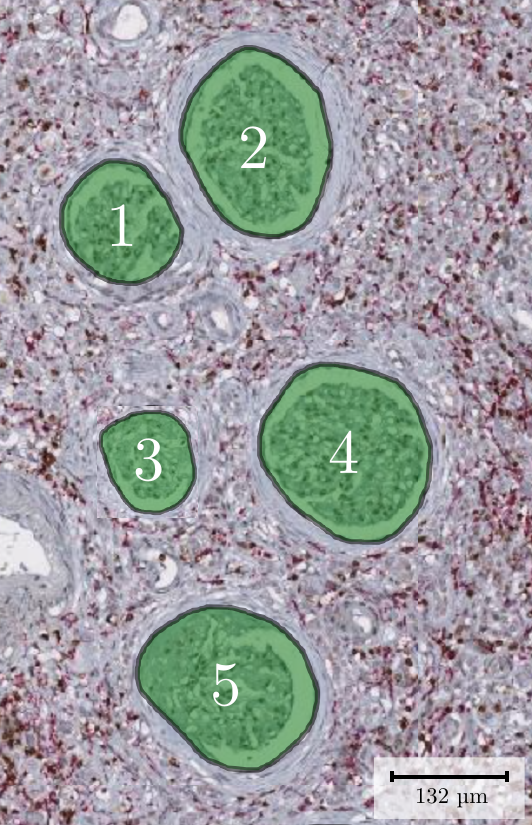}
%        \label{}
    \end{subfigure}
	\caption{Subsamples of the four consecutive kidney nephrectomy WSIs of Patient 1 used in this study. From left to right row: CD3-CD68, CD3-CD163, CD3-CD206, CD3-MS4A4A. The number in the bottom row shows the glomeruli matching ground truth.}
	\label{Fig: example multistained WSI study crop}
\end{figure*}

Whole slide images were acquired using an Aperio AT2 scanner at $40{\times}\!$ magnification (a resolution of \SI[per-mode=fraction]{0.253}{\micro\metre\per{pixel}}).
All the healthy and sclerotic glomeruli in each WSI were annotated by outlining them using Cytomine \cite{maree2016collaborative} and validated by pathology experts. The number of glomeruli for each patient and in each staining is summarised in Table \ref{tbl:glomerulinumbers}. For technical reasons (most likely due to uneven tissue fixation), staining artefacts occurred in patient 2, that resulted in the need for manual removal of some  areas. As the affected tissue was removed from the evaluation, Table \ref{tbl:glomerulinumbers} reports both the number of glomeruli including the ignored tissue (in parentheses) and the final corrected results. The WSIs and annotations of patient 1 are shown in Fig.\ \ref{Fig: example multistained WSI study} and larger scale crops in Fig.\ \ref{Fig: example multistained WSI study crop}.

To validate the matching algorithm\footnote{The code is publicly available from: \url{https://gitlab.in2p3.fr/odyssee.merveille/glomeruli_matching-cmpb-2021.git}.\color{black}}, approximately $270$ glomeruli were manually associated with each other between the four slides of patient 1 (including $220$ that exist within all four slides), and approximately $185$ glomeruli in patient 2 (including $169$ that exist within all four slides).

%%%%%%%%%%%%%%%%%%%%%%%%%%%%%%%%%%%%%%%%%%%%%%%%%%%%%%%%%%%%%%%%%%
%%%%%%%%%%%%%%%% Automatic analysis pipeline %%%%%%%%%%%%%%%%%%%%%
%%%%%%%%%%%%%%%%%%%%%%%%%%%%%%%%%%%%%%%%%%%%%%%%%%%%%%%%%%%%%%%%%%

\begin{table}[t]
    \centering
    \caption{Number of glomeruli per patient and stain. The numbers in brackets for patient $2$ are the number of glomeruli before removing those situated in areas affected by staining irregularities, please refer to the text for more information.}
    \footnotesize{
    \begin{tabular}{ r  c  c  c  c }
    \toprule
     Patient & CD3-CD68 & CD3-CD163 & CD3-CD206 & CD3-MS4A4A\\
     \midrule
    1 & $445$ & $482$ & $480$ & $470$\\
    2 & $185$ ($271$) & $173$ ($255$)  & $180$ ($267$) & $176$ ($253$)\\
    3 & $135$ & $128$ & $130$ & $122$\\
    4 & $285$ & $254$ & $274$ & $244$ \\
    \midrule
    Total & $1050$ & $1037$ & $1064$ & $1012$\\
    \bottomrule
    \end{tabular}
    }
    \label{tbl:glomerulinumbers}
\end{table}

%%%%%%%%%%%%%%%%%%%%%%%%%%%%%%%%%%%%%%%%%%%%%%%%%%%%%%%%%%%%%%%%%%
%%%%%%%%%%%%%%%%%%%%%%%%%%%% Results %%%%%%%%%%%%%%%%%%%%%%%%%%%%%
%%%%%%%%%%%%%%%%%%%%%%%%%%%%%%%%%%%%%%%%%%%%%%%%%%%%%%%%%%%%%%%%%%

\section{Results}
\label{sec:results}

In this section, the proposed matching algorithm is validated and then the results of the full pipeline are presented.

The following metrics are used to evaluate matching performance: Sensitivity ($S = \tfrac{\text{TP}}{\text{TP}+\text{FN}}$), Precision ($P = \tfrac{\text{TP}}{\text{TP}+\text{FP}}$), and Specificity ($SP = \tfrac{\text{TN}}{\text{TN}+\text{FP}}$) and Negative Predictive Value ($\text{NPV} = \tfrac{\text{TN}}{\text{TN}+\text{FN}}$) to account for the possibility of false positive---a centroid incorrectly associated to another---and true negative associations---unpaired centroids not associated with another correctly. The values of TP, FP, and FN were measured in terms of associations, such that a TP is a correct association, an FP is an incorrect association, and an FN occurs when an association is missed.

\subsection{Validation on Glomeruli Ground-Truth Segmentation}

\begin{figure*}[t]
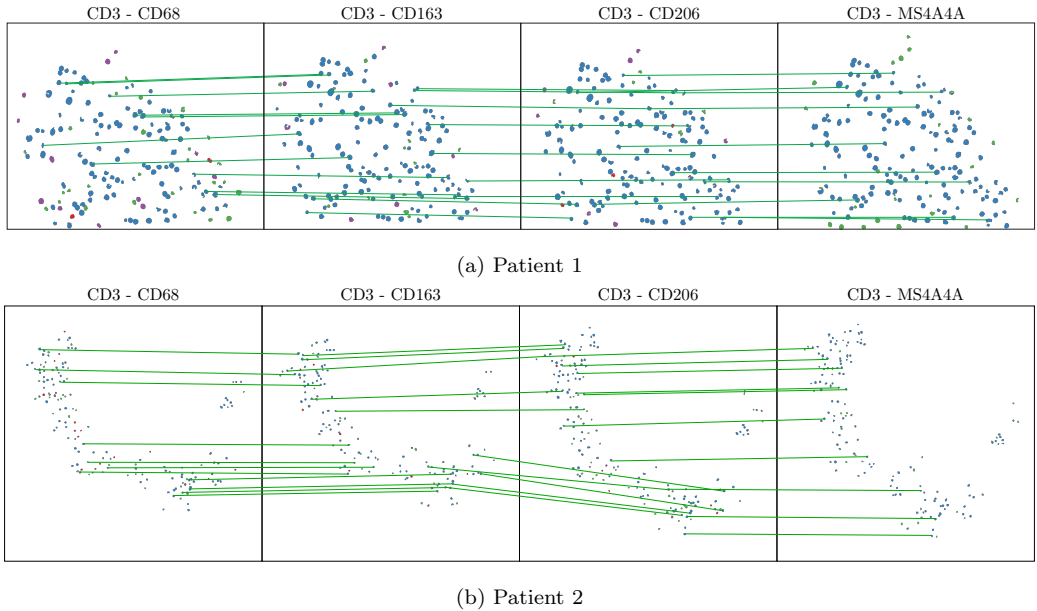

	\centering
	\begin{subfigure}[t]{\linewidth}
        \includegraphics[width=\linewidth]{./images/p1_matching.pdf}
        \caption{Patient 1}
%        \label{fig:gt_matching_p1}
    \end{subfigure}
	\hspace{0.05cm}
	\begin{subfigure}[t]{\linewidth}
        \includegraphics[width=\linewidth]{./images/p2_matching.pdf}
        \caption{Patient 2}
%        \label{fig:gt_matching_p2}
    \end{subfigure}
	\caption{A random subset of the glomeruli matchings between the four WSIs, in which TN matched glomeruli are in green, TP in blue (the green line represents the correct associations between WSI), FN in purple, and FP in red.}
	\label{fig:gt_matching}
\end{figure*}

\begin{table*}[h!]
    \centering
    \caption{Matching performance on ground-truth (GT) vs. on segmentation (Segm) with $d_{\text{match}} = 300 $ and $N = 4$  ($S$ = Sensitivity, $SP$ = Specificity, $P$ = Precision, $NPV$ = Negative Predictive Value).\color{black}}
    \begin{subfigure}[t]{1\textwidth}
    \centering
    \footnotesize
    \begin{tabular}{r c c c c c c c c}
    \toprule
    \multirow{2}{*}{Stain Pair} & \multicolumn{2}{c}{$S$} & \multicolumn{2}{c}{$SP$} & \multicolumn{2}{c}{$P$} & \multicolumn{2}{c}{$NPV$}\\
     & GT & Segm & GT & Segm & GT & Segm & GT & Segm\\
    \midrule
    CD3-CD68 --- CD3-CD163 & $93\%$ & $86\%$ & $90\%$ & $89\%$  & $99\%$ & $97\%$  & $43\%$ & $58\%$ \\
    CD3-CD163 --- CD3-CD206 & $98\%$ & $94\%$ & $100\%$ & $100\%$ & $100\%$ & $100\%$  & $75\%$ & $55\%$\\
    CD3-CD206 --- CD3-MS4A4A & $96\%$ & $93\%$ & $100\%$ & $57\%$ & $100\%$ & $98\%$  & $43\%$ & $27\%$\\
    \bottomrule
    \end{tabular}
    \caption*{Patient 1}
    \end{subfigure}\\
    \begin{subfigure}[t]{1\textwidth}
    \centering
    \footnotesize
    \begin{tabular}{ r c c c c c c c c}
    \toprule
    \multirow{2}{*}{Stain Pair} & \multicolumn{2}{c}{$S$} & \multicolumn{2}{c}{$SP$} & \multicolumn{2}{c}{$P$} & \multicolumn{2}{c}{$NPV$}\\
     & GT & Segm & GT & Segm & GT & Segm & GT & Segm\\
    \midrule
    CD3-CD68 --- CD3-CD163 & $95\%$ & $92\%$ & $94\%$ & $74\%$  & $99\%$ & $90\%$  & $80\%$ & $76\%$ \\
    CD3-CD163 --- CD3-CD206 & $98\%$ & $93\%$ & $100\%$ & $67\%$ & $100\%$ & $94\%$  & $84\%$ & $57\%$\\
    CD3-CD206 --- CD3-MS4A4A & $94\%$ & $95\%$ & $95\%$ & $85\%$ & $99\%$ & $95\%$  & $70\%$ & $82\%$\\
    \bottomrule
    \end{tabular}
    \caption*{Patient 2}
    \end{subfigure}
    \label{tbl:matching_perf}
\end{table*}

\begin{figure}[t]
	\centering
	\includegraphics[width=0.6\linewidth]{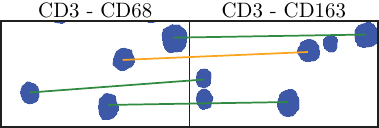}
	\caption{False positive matching occurring in patient 1 between CD3-CD68 and CD3-CD163 (in orange) when applying the matching algorithm to the ground-truth segmentation. The correctness of this association is debatable even for experts}
	\label{fig:gt_matching_p1_fp}
\end{figure}

The matching algorithm was first validated independently of the pipeline, more specifically of possible segmentation errors, by matching the glomeruli of the nephrectomy dataset obtained by manual segmentation (Fig.\ \ref{Fig: example multistained WSI study}). The results of these experiments are shown in Table \ref{tbl:matching_perf} and Fig.\ \ref{fig:gt_matching}.

The matching algorithm has little trouble finding correct associations in all but a very few cases as shown by the very high sensitivity and precision. Moreover the false associations are usually understandable as they concern glomeruli that are in close proximity, and the ground truth matching was problematic even for experts (see Fig.\ \ref{fig:gt_matching_p1_fp}).
Most of the errors concern false detections, as shown by the NPV score. The variance of NPV is high as it is computed on a small number of samples (one more false negative association will decrease the NPV by a few tens of percent). When a clear association cannot be found, the algorithm tends not to match the glomerulus, which is a desired behaviour for the discussed applications since it ensures that associations are reliable and will not bias further statistics that could be built upon them.

\subsection{Matching on Glomeruli Segmentation}
\label{sec:resseg}

The stainings in this study present similar visual characteristics, see Fig.\ \ref{Fig: example multistained WSI study crop}, which lends to training one `multi-stain' U-Net by combining the training sets of each stain and applying the same network to all stains. To better utilise the limited amount of data, one network was trained for each patient in a leave-one-out fashion, such that the segmentor for patient 1 was trained using data from patients 2, 3, and 4; and the one for patient 2 was trained using the data derived from patients 1, 3, and 4. The training set comprised patches centred on all glomeruli from the training patients and seven times the number of tissue patches (to account for the variance observed in non-glomeruli tissue), $20\%$ of this data was reserved for validation.
\color{black}

The segmentation performance of this approach is described in Table \ref{tbl:segmentation}.

The centroids of each detected glomerulus were extracted to form the sets $G$ and $H$, which are the input to the matching algorithm. Pairwise matching was then performed on each consecutive image to determine the associations between all WSIs. The results are presented in Table \ref{tbl:matching_perf}.
The sensitivity and precision are still very high compared to the ground-truth baseline, which demonstrates the algorithm's detection robustness. Specificity sometimes drops significantly and these drops are not accompanied by a significant drop in sensitivity or precision. This is explained by the very small number of negative matchings in these stainings, and each single false positive match yields a large specificity drop. This behaviour is not problematic in a global scale as the number of false positive matches remains very low.

\begin{table*}[b]
    \centering
    \caption{Segmentation performance of the detection algorithm based on pixels (average $F_1$ score of five repetitions, with standard deviations in parentheses).}
    \footnotesize
    \begin{tabular}{ r  c  c  c  c  c }
    \toprule
     Patient & CD3-CD68 & CD3-CD163 & CD3-CD206 & CD3-MS4A4A & Overall\\
     \midrule
    1 & $0.803$ ($0.010$) & $0.801$ ($0.020$) & $0.822$ ($0.015$) & $0.818$ ($0.014$) & $0.811$ ($0.014$)\\
    2 & $0.860$ ($0.003$) & $0.863$ ($0.005$) & $0.859$ ($0.015$) & $0.865$ ($0.011$) & $0.862$ ($0.003$)\\
    \bottomrule
    \end{tabular}
    \label{tbl:segmentation}
\end{table*}

\subsection{Multi-WSI Analysis}

\begin{figure}[h!]
	\centering
	\begin{subfigure}[c]{0.45\linewidth}
	\begin{subfigure}[c]{0.45\linewidth}
        \includegraphics[width=\linewidth,frame]{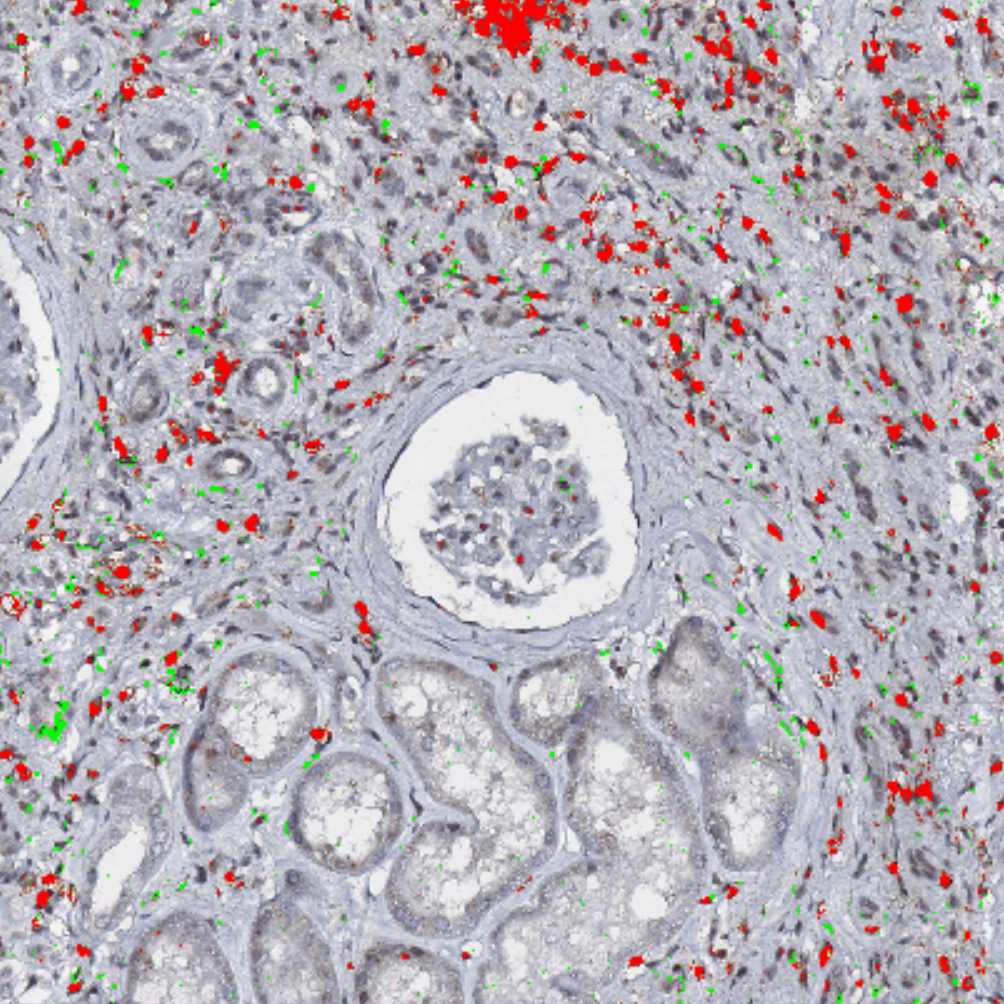}
        \caption{}
    \end{subfigure}%
	\hspace{0.05cm}
	\begin{subfigure}[c]{0.45\linewidth}
        \includegraphics[width=\linewidth,frame]{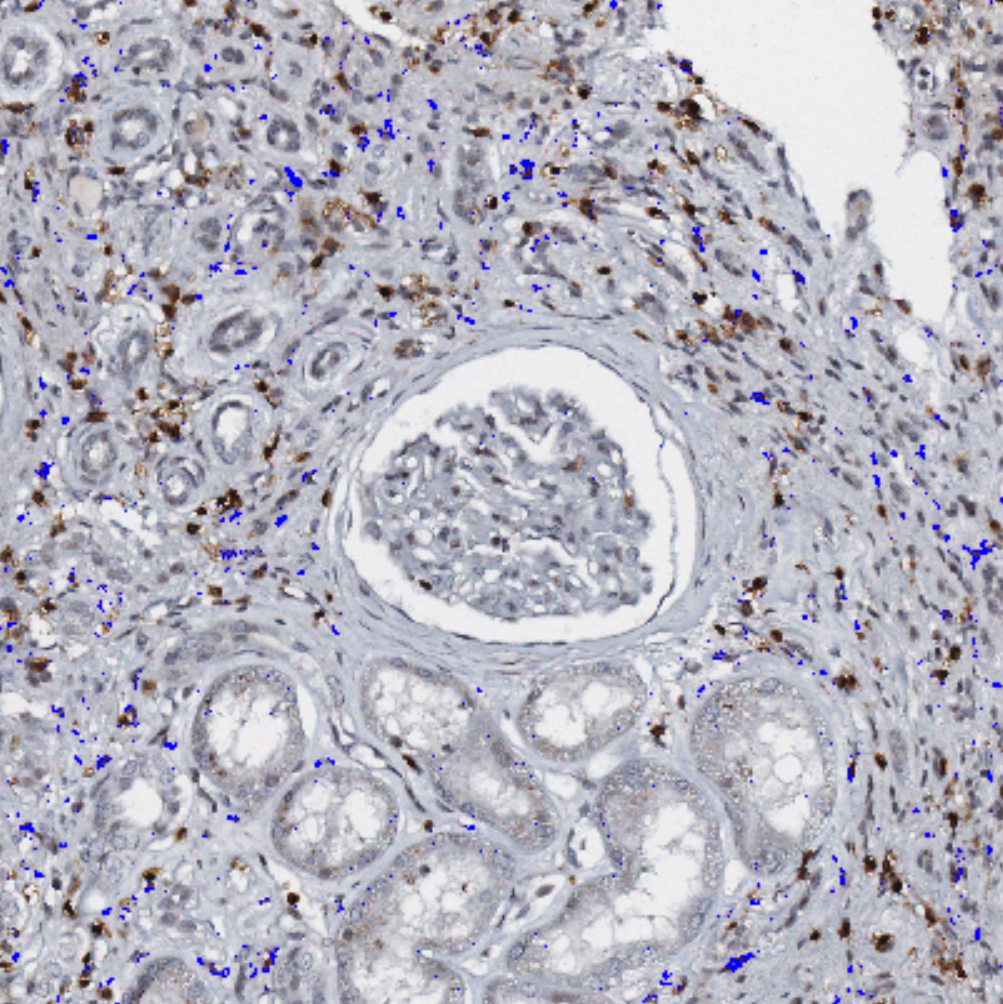}
        \caption{}
    \end{subfigure}%
    \\
	\begin{subfigure}[c]{0.45\linewidth}
        \includegraphics[width=\linewidth,frame]{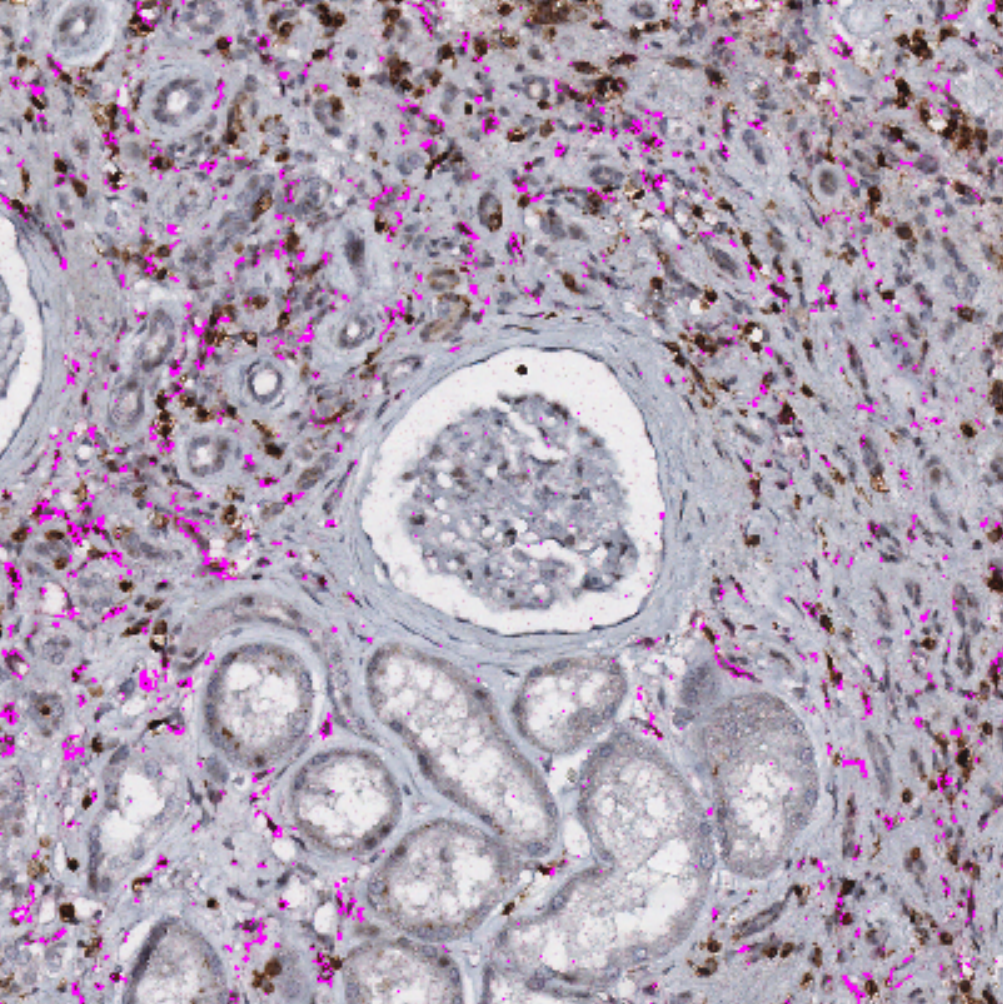}
        \caption{}
    \end{subfigure}%
    \hspace{0.05cm}
	\begin{subfigure}[c]{0.45\linewidth}
        \includegraphics[width=\linewidth,frame]{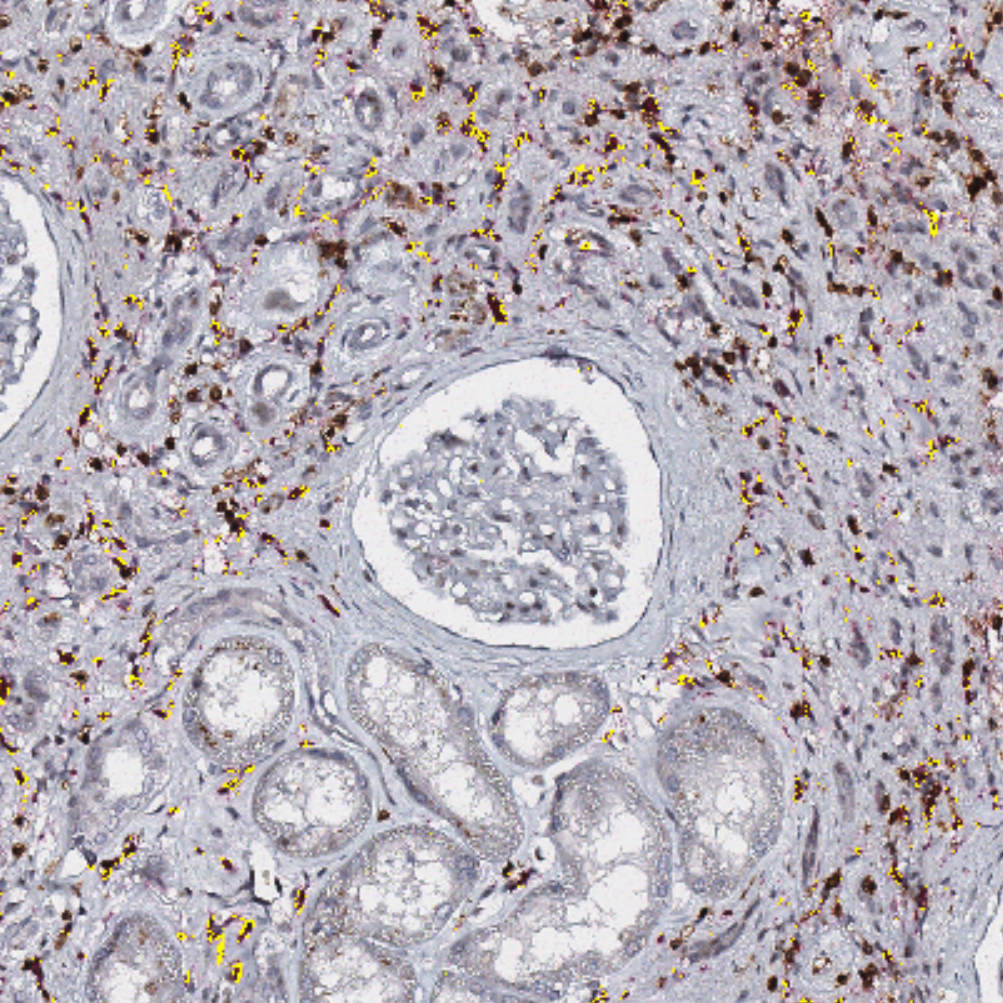}
        \caption{}
    \end{subfigure}%
    \end{subfigure}%
    \hspace{0.05cm}
	\begin{subfigure}[c]{0.45\linewidth}
        \includegraphics[width=\linewidth,frame]{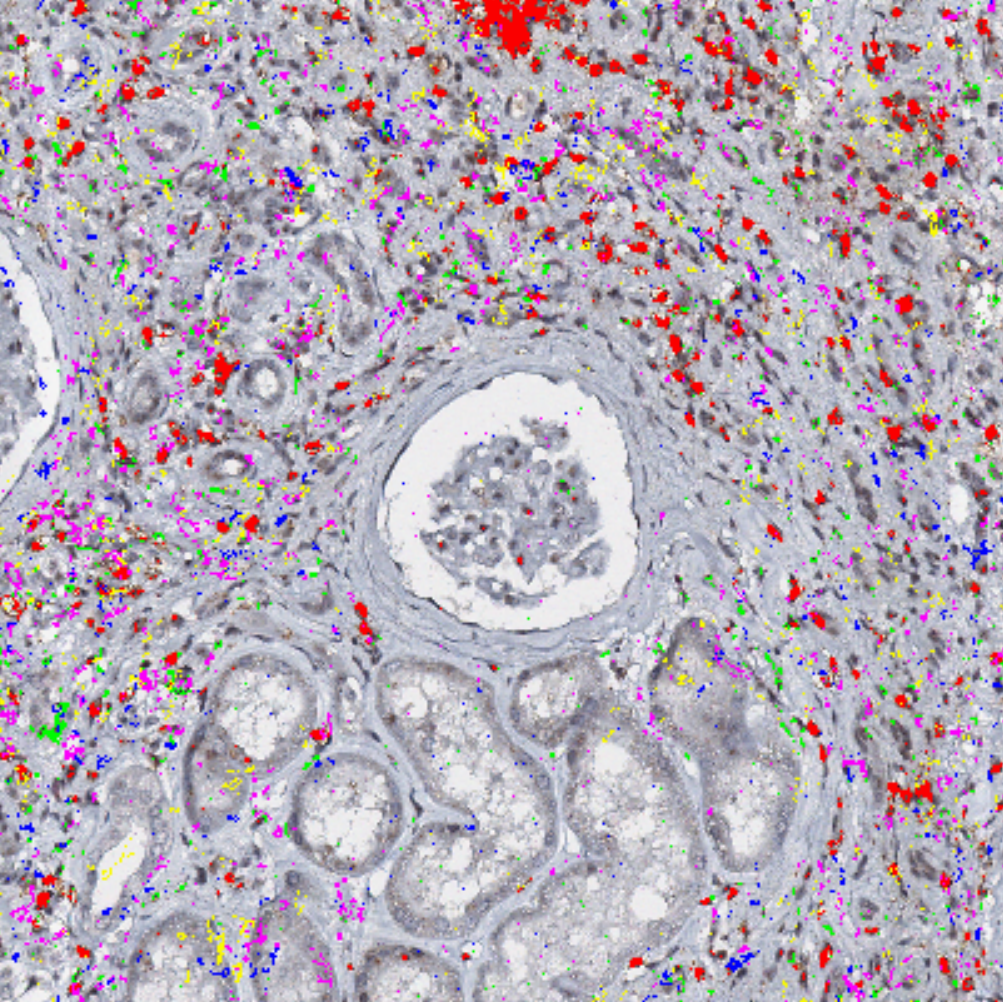}
        \caption{}
    \end{subfigure}%
	\caption{Illustrating the segmentation of different cell types around the same glomerulus through $4$ consecutive slides. (a) T cells (red) and M0 macrophages (green) in CD3-CD38 WSI, (b) M2 macrophages (blue) in CD3-CD163 WSI, (c) M2 macrophages (pink) in CD3-CD206 WSI, (d) M2 macrophages (yellow) in CD3-MS4A4A WSI, (e) superimposition of all cell types in CD3-CD68 WSI. Note that each subtype of macrophage is present in only one of the consecutive slides.}
	\label{Fig: features illustration}
\end{figure}

At this stage of the pipeline, it is possible to register each matched glomerulus and its surrounding, allowing the superimposition of the segmentations from each consecutive WSI. Fig.\ \ref{Fig: features illustration} shows the result of this for a glomerulus of Patient 1. It should be emphasised that this image illustrates something that would not be possible with conventional staining techniques: a glomerulus with the combined information requiring five separate stainings (M0 macrophages, 3 different polarisations of M2 macrophages, and T-cells). \color{black} 
The proposed framework therefore enables features of the tissue to be extracted that were previously not possible. Both Multi-WSI and Intra-WSI features can be computed and used for diagnosis and research purposes. 

\begin{figure}[h!]
	\centering
	\begin{subfigure}[t]{0.45\linewidth}
        \includegraphics[width=\linewidth,frame]{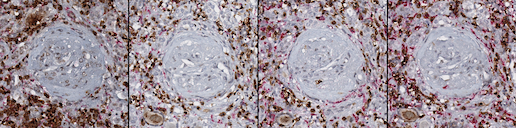}
        \caption{Score $0.000$}
%        \label{fig:gt_matching_p1}
    \end{subfigure}
    \begin{subfigure}[t]{0.45\linewidth}
        \includegraphics[width=\linewidth,frame]{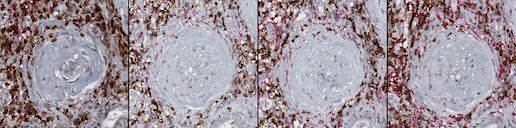}
        \caption{Score $0.035$}
%        \label{fig:gt_matching_p1}
    \end{subfigure}
    \\
    \begin{subfigure}[t]{0.45\linewidth}
        \includegraphics[width=\linewidth,frame]{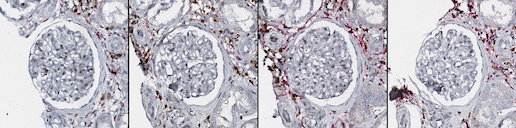}
        \caption{Score $0.49$}
%        \label{fig:gt_matching_p1}
    \end{subfigure}
    \\
    \begin{subfigure}[t]{0.45\linewidth}
        \includegraphics[width=\linewidth,frame]{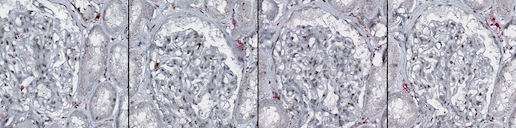}
        \caption{Score $0.961$}
%        \label{fig:gt_matching_p1}
    \end{subfigure}
    \begin{subfigure}[t]{0.45\linewidth}
        \includegraphics[width=\linewidth,frame]{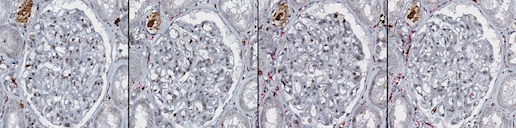}
        \caption{Score $1.000$}
%        \label{fig:gt_matching_p1}
    \end{subfigure}
    \\
    \begin{subfigure}[t]{0.5\linewidth}
        \includegraphics[width=\linewidth]{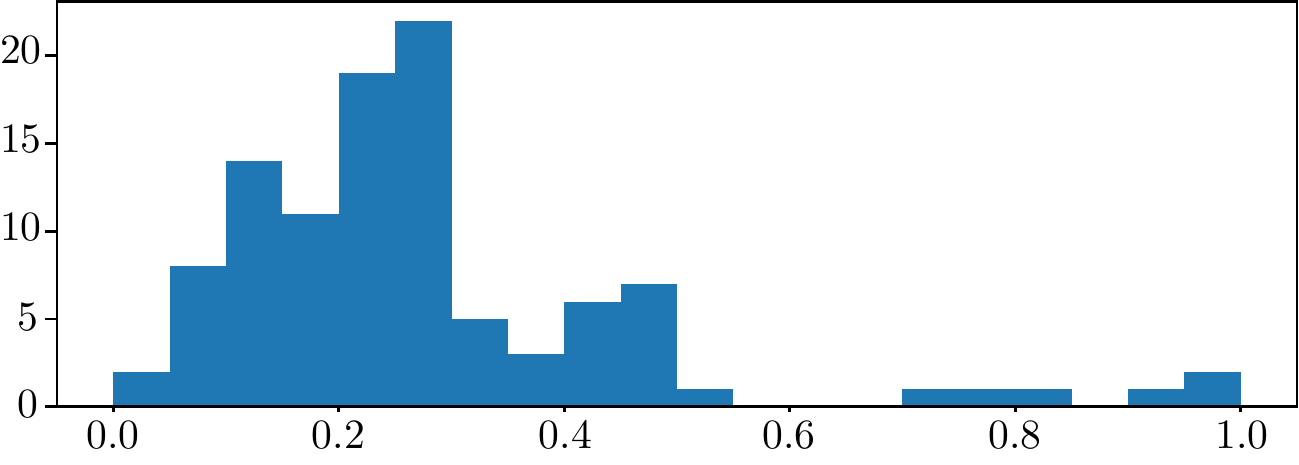}
        \caption{Score Histogram}
%        \label{fig:gt_matching_p1}
    \end{subfigure}
	\caption{Patient 1 matched glomerulus ranking: a) and b) bottom two, c) middle score, and d) and e) top two. Scores have been normalised to between 0 and 1. The images for each glomerulus are (from left to right): CD3-CD68, CD3-CD163, CD3-CD206, CD3-MS4A4A.}
	\label{fig:p1_glom_rank}
\end{figure}

\begin{figure}[h!]
	\centering
	\begin{subfigure}[t]{0.45\linewidth}
        \includegraphics[width=\linewidth,frame]{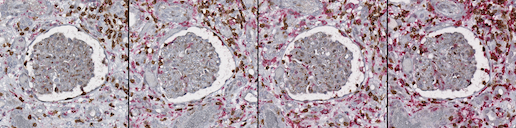}
        \caption{Score $0.000$}
%        \label{fig:gt_matching_p1}
    \end{subfigure}
    \begin{subfigure}[t]{0.45\linewidth}
        \includegraphics[width=\linewidth,frame]{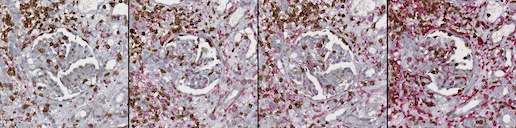}
        \caption{Score $0.035$}
%        \label{fig:gt_matching_p1}
    \end{subfigure}
    \\
    \begin{subfigure}[t]{0.45\linewidth}
        \includegraphics[width=\linewidth,frame]{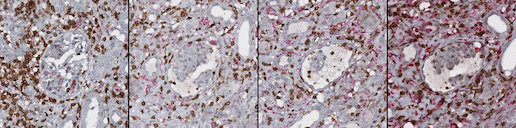}
        \caption{Score $0.506$}
%        \label{fig:gt_matching_p1}
    \end{subfigure}
    \\
    \begin{subfigure}[t]{0.45\linewidth}
        \includegraphics[width=\linewidth,frame]{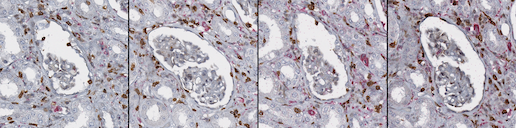}
        \caption{Score $0.961$}
%        \label{fig:gt_matching_p1}
    \end{subfigure}
    \begin{subfigure}[t]{0.45\linewidth}
        \includegraphics[width=\linewidth,frame]{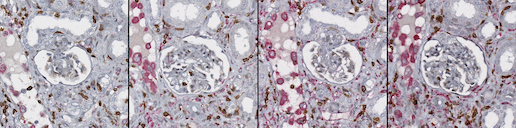}
        \caption{Score $1.000$}
%        \label{fig:gt_matching_p1}
    \end{subfigure}
    \\
    \begin{subfigure}[t]{0.5\linewidth}
        \includegraphics[width=\linewidth]{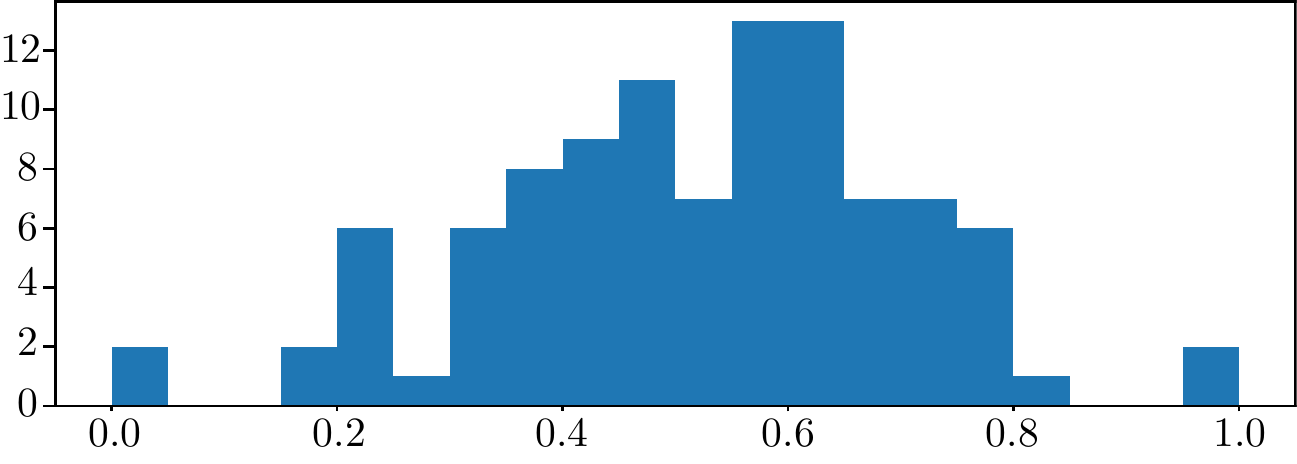}
        \caption{Score Histogram}
%        \label{fig:gt_matching_p1}
    \end{subfigure}
	\caption{Patient 2 matched glomerulus ranking: a) and b) bottom two, c) middle score, and d) and e) top two. Scores have been normalised to between 0 and 1. The images for each glomerulus are (from left to right): CD3-CD68, CD3-CD163, CD3-CD206, CD3-MS4A4A.}
	\label{fig:p2_glom_rank}
\end{figure}

\begin{figure}[h!]
	\centering
\includegraphics[width=0.5\linewidth,frame]{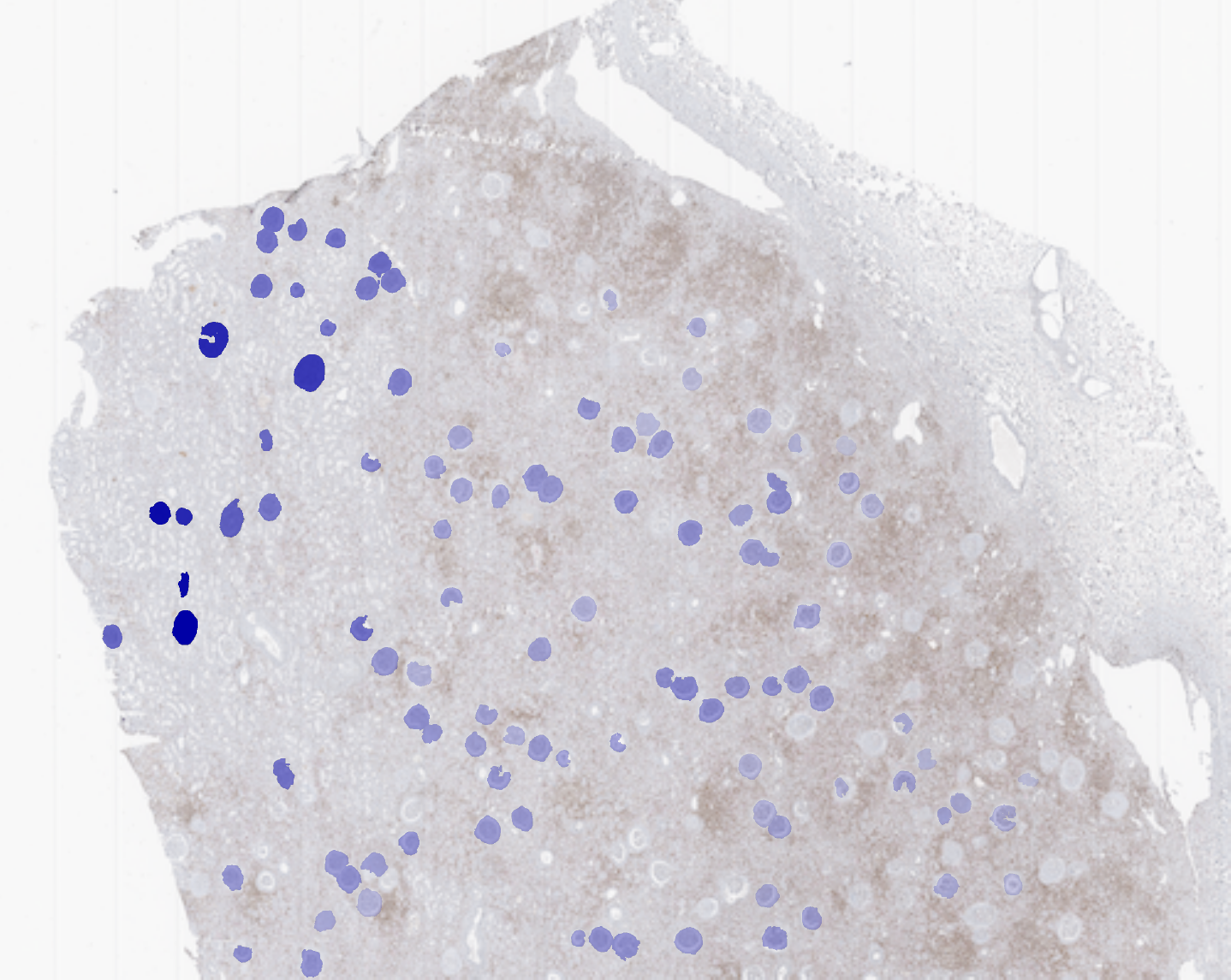}
	\caption{Patient 1 glomerulus ranking spatial distribution in which transparency is proportional to rank (i.e.\ more transparent blue regions are ranked lower).}
	\label{fig:p1_glom_rank_spatial}
\end{figure}

To illustrate the type of analysis that can now be achieved, a `toy' unsupervised glomerulus ranking score based on the 
information derived from each stain has been developed. The Multi-WSI and Intra-WSI features can also be used to perform other analyses such as clustering the glomeruli to expose different groupings, training a supervised classification model using the features, or performing in-depth statistical analyses.

To create this score, a matrix $\mathbf{F}=\mathbb{R}^{|M| \times 19}$, where $|M|$ is the number of matched glomeruli, is constructed with the following $19$ features extracted from a neighbourhood of size $\SI{258}{\micro\meter}$ centred on each matched glomerulus, these $19$ features comprised the following $12$ multi-stain features:
\begin{itemize}[noitemsep,topsep=0pt]
\item[-] mean M0 macrophage (CD68), M2 macrophage (CD163), M2 macrophage (CD206), M2 macrophage (MS4A4A) density inside glomeruli;
\item[-] mean M0 macrophage (CD68), M2 macrophage (CD163), M2 macrophage (CD206), M2 macrophage (MS4A4A) density outside glomeruli;
\item[-] mean distance from M0 macrophage (CD68), M2 macrophage (CD163), M2 macrophage (CD206), M2 macrophage (MS4A4A) to glomeruli;
\end{itemize}
and $7$ intra-stain features:
\begin{itemize}[noitemsep,topsep=0pt]
\item[-] mean T-cells (CD3) density inside glomeruli over all stainings;
\item[-] mean T-cells (CD3) density outside glomeruli over all stainings;
\item[-] mean distance from M0 macrophage (CD68), M2 macrophage (CD163), M2 macrophage (CD206), M2 macrophage (MS4A4A) to T-cells (CD3);
\item[-] mean distance from T-cells (CD3) to glomeruli over all stainings.
\end{itemize}
It should be \color{red} emphasised\color{black}, that this fusion of information (combination of cell features from different stainings) would not be possible without the proposed framework. \color{black} 
Principal component analysis is used to extract the first principal component that explains the most correlated variance between the features, and the glomerulus ranking score is the value of this component.

The glomeruli associated with the two highest scores, the two lowest scores, and the one falling in the middle of the scale are presented in Fig.\ \ref{fig:p1_glom_rank} for patient 1 and in Fig.\ \ref{fig:p2_glom_rank} for patient 2. Histograms of the score distributions for each patient are also presented and these show that patient 1's glomeruli are skewed towards the lower ranked end of the scale whereas patient 2's glomeruli are distributed around intermediate scores. 

Visual inspection suggests that the scores are associated with the severity of glomerular sclerosis: in general, low scores are associated with more severe glomerulosclerosis when compared to high scores. This may indicate a role of the inflammatory micro-environment surrounding glomeruli for pro-fibrotic changes. This is interesting, because the score is not directly reflecting cell numbers but instead is driven by the $19$ features reflecting density and complex neighbourhood relationships of immune cells. In addition, the analysis provides information on the spatial distribution of glomeruli in their micro-environment, e.g.\ for patient 1, the more severely affected glomeruli with low scores tend to be localised more superficially (close to the renal capsule) and in association with dense inflammatory infiltrates. In contrast, largely normal glomeruli with high scores are clustered in moderately inflamed areas (Fig.\ \ref{fig:p1_glom_rank_spatial}).

This is a very naive demonstration of this pipeline to demonstrate its viability and usefulness. The next stage of research will be to develop more complex scores based on multi-stain features that are clinically relevant in order to further study the glomeruli microenvironment in relation to IFTA and glomerulosclerosis.

%%%%%%%%%%%%%%%%%%%%%%%%%%%%%%%%%%%%%%%%%%%%%%%%%%%%%%%%%%%%%%%%%%%%%%%%%%%%%%%%%%%%%%%%%%%%%%%%% Conclusion %%%%%%%%%%%%%%%%%%%%%%%%%%%%%
%%%%%%%%%%%%%%%%%%%%%%%%%%%%%%%%%%%%%%%%%%%%%%%%%%%%%%%%%%%%%%%%%%%%%

\section{Conclusions}

In summary, this article has presented a novel framework for the study of tissue micro-environment of renal glomeruli across multiple WSIs that allows their comprehensive evaluation without technically challenging multiplexing, by integrating multiple staining modalities in consecutive tissue sections. The framework involves approximate rigid registration and segmenting glomeruli and cells in each WSI, which can be achieved using standard algorithms, then matching them across the WSIs to integrate the information contained within each. The result of this can then be used to perform analyses on the glomeruli and surrounding tissue. 

The proposed framework is generic and independent of the presented use-cases. It can be used for the analysis of the micro-environment surrounding other large structures, under the assumption that such structures are large enough to exist across multiple WSIs and can be segmented (either manually or automatically). Furthermore, it is independent of the segmentation algorithm used and can therefore be applied to a variety of biomedical research questions beyond transplantation medicine, for example immuno-oncology and other scientific fields working with biopsy samples.

In the future, this approach could support the diagnosis of renal grafts by time-efficient quantification and evaluation of glomeruli (e.g.\ \color{red} the distinction between normal and altered glomeruli of different underlying pathological processes and severity grades\color{black}) 
and precise number and localisation of infiltrating leukocytes (e.g.\ glomerulitis according to the internationally used Banff classification for renal grafts \cite{Haas18}). Counting glomeruli with the described methods could also be performed for 3D reconstruction (research purposes) in consecutive tissue slides and thus enable an estimation of glomeruli numbers in the whole kidney; reduced renal allograft survival \cite{Mackenzie96}, hypertension and the risk of chronic kidney disease \cite{Tsuboi14} are associated with low glomeruli number.

\section*{Acknowledgement}
This work was supported by ERACoSysMed project ``SysMIFTA'', co-funded by EU H2020 and the national funding agencies German Ministry of Education and Research (BMBF) project management PTJ (FKZ: 031L-0085A), and the the e:Med project SYSIMIT, project management DLR (FKZ: 01ZX1608B), and Agence National de la Recherche (ANR), project number ANR-15-CMED-0004.

The authors thank Nvidia Corporation for donating a Quadro P6000 GPU, the \emph{Centre de Calcul de l'Université de Strasbourg} for access to the GPUs used for this research, and Nicole Krönke for her excellent technical assistance in data preparation.

\clearpage
\setcounter{page}{1}
\setcounter{section}{1}
\setcounter{table}{0}
\setcounter{figure}{0}
\section*{Supplementary Material}

\subsection{Glomeruli and Cell Segmentation}
\label{sec:gcseg}
\subsubsection{Glomeruli Segmentation}
\label{SubSec: Glomeruli Segmentation}

Two approaches can be taken to segment the glomeruli slices in all stainings: develop a segmentation model for each staining \cite{Samsi12,Gadermayr16,Gadermayr16b, Maree16,Gallego18,de2018automatic,Kannan19}, or a stain invariant/multi-stain segmentation model \cite{Lampert19}.

Computer vision approaches such as perceptual organisation \cite{Samsi12}, histogram of gradients \cite{Gadermayr16}, colour profiles \cite{Gadermayr16}, local binary patterns \cite{Gadermayr16b, Simon18}, and combinations of approaches \cite{Maree16} integrate background knowledge into the task. Nevertheless, there is no general consensus on the type of features to extract and so data driven approaches have gained in popularity. Most recently, deep learning approaches \cite{Kannan19, Gallego18, de2018automatic} have become the de-facto standard for segmentation due to their state-of-the-art performance, however, being data driven they require a large amount of training data. To overcome this, pretrained networks such as GoogleNet, AlexNet \cite{Gallego18}, and VGG16 \cite{Marsh18} can be used.

The proposed matching framework is agnostic to the segmentation algorithm used. In the demonstrated application, segmentation is performed using a U-Net \cite{Ronneberger15} as it has been proven to be successful in biomedical imaging \cite{litjens2017survey}, in particular in glomeruli detection \cite{de2018automatic,Lampert19}. 
Nevertheless, any of the above-mentioned (or other) approaches can be used to produce the segmentation for this stage of the pipeline in case of difficulty applying the U-Net.

Glomeruli segmentation is framed as a two classes problem: \emph{glomeruli} and \emph{tissue}. The slide background (non-tissue) is manually removed from consideration. The input to the network are patches centred on glomeruli (as defined by the ground truth, see Sec.\ 
\ref{sec: data}), and those that do not contain a glomerulus, randomly sampled.

The U-Net was implemented as described in the original article \cite{Ronneberger15} using the cross entropy loss and trained for multi-stain segmentation (see Section \ref{sec:resseg})\footnote{Because of the relatively small amount of training data in the experiments presented in Section \ref{sec:results}, and the large variance observed between the stainings and characteristics of each patient, the U-Net used upsampling instead of transposed convolution to reduce the number of learnable parameters.}. The following parameter values were used: batch size of $8$, learning rate of $0.0001$, $60$ epochs, and the network that achieves the lowest validation loss is kept. The input patch size is $508 \times 508$ pixels, which is sufficient to contain a glomerulus at a resolution of $\SI[per-mode=fraction]{0.506}{\micro\metre\per{pixel}}$.

The following data augmentation is performed with an independent probability of $0.5$:
\begin{description}[noitemsep,topsep=0pt,leftmargin=0pt]
\item[elastic deformation:] using the parameters $\sigma = 10$, $\alpha = 100$;
\item[affine:] random rotation sampled from the interval $\interval{\ang{0}}{\ang{180}}$, random shift sampled from $\interval{-205}{205}$ pixels, random magnification sampled from $\interval{0.8}{1.2}$, and horizontal/vertical flip;
\item[noise:] additive Gaussian noise with $\sigma \in \interval{0}{2.55}$;
\item[blur:] Gaussian filter with $\sigma \in \interval{0}{1}$;
\item[brightness] enhance with a factor sampled from $\interval{0.9}{1.1}$;
\item[colour] enhance with a factor sampled from  $\interval{0.9}{1.1}$;
\item[contrast] enhance with a factor sampled from $\interval{0.9}{1.1}$.
\end{description}
These values were chosen to produce realistic images. All samples are standardised to have a minimum value of 0 and maximum of 1 and normalised by the mean and standard deviation of the training set.

During application, the U-Net was applied using the `overlap-tile strategy' \cite{Ronneberger15}. Furthermore, the output of the U-Net was postprocessed by removing the smallest connected components and closing small holes.

\subsubsection{Cell Segmentation}

Our dataset is composed of $4$ double-stained consecutive WSIs for each patient, each staining highlighting different cell types (see Sec.\ \ref{sec: data}). In total, $5$ different cell types are highlighted: T cells (CD3) and $4$ different types of M2-like macrophages (CD68, CD163, CD206 and MS4A4A).
The goal of this step is to segment each cell type resulting in $5$ binary images that will be used to compute features.

The image resulting from the digitisation of a WSI is a mixture of the signals from two stains (e.g.\ CD3 and CD68) and the counter-stain (e.g.\ haematoxylin).
The classic method to unmix the stains from an RGB image was proposed by Ruifrok et al.\ and called colour deconvolution \cite{Ruifrok01}. This method transforms the RGB channels of the WSI into optical densities of each staining that are linearly related to their concentrations in the tissue. Once each slide is unmixed, a simple thresholding of the channels of interest is enough to segment the structures targeted by the main stain.

Colour deconvolution requires a predetermined stain vector for each staining that represent the proportion of optical densities of this staining in each RGB channel. In this work, the stain vectors for each staining were measured from the dataset, however, unsupervised methods have been proposed for situations in which stain vector measurement is not an option. These methods are based on singular value decomposition \cite{Macenko09}, blind deconvolution \cite{Gavrilovic13}, dictionary learning \cite{Vahadane16},  multilayer perceptron networks \cite{Wemmert13} or non-negative matrix factorisation \cite{Rabinovich14}.
\color{black}

\subsection{Glomeruli Matching Synthetic Validation}

\begin{figure}[!ht]
	\centering
	\begin{subfigure}[t]{0.45\linewidth}
        \includegraphics[width=\linewidth]{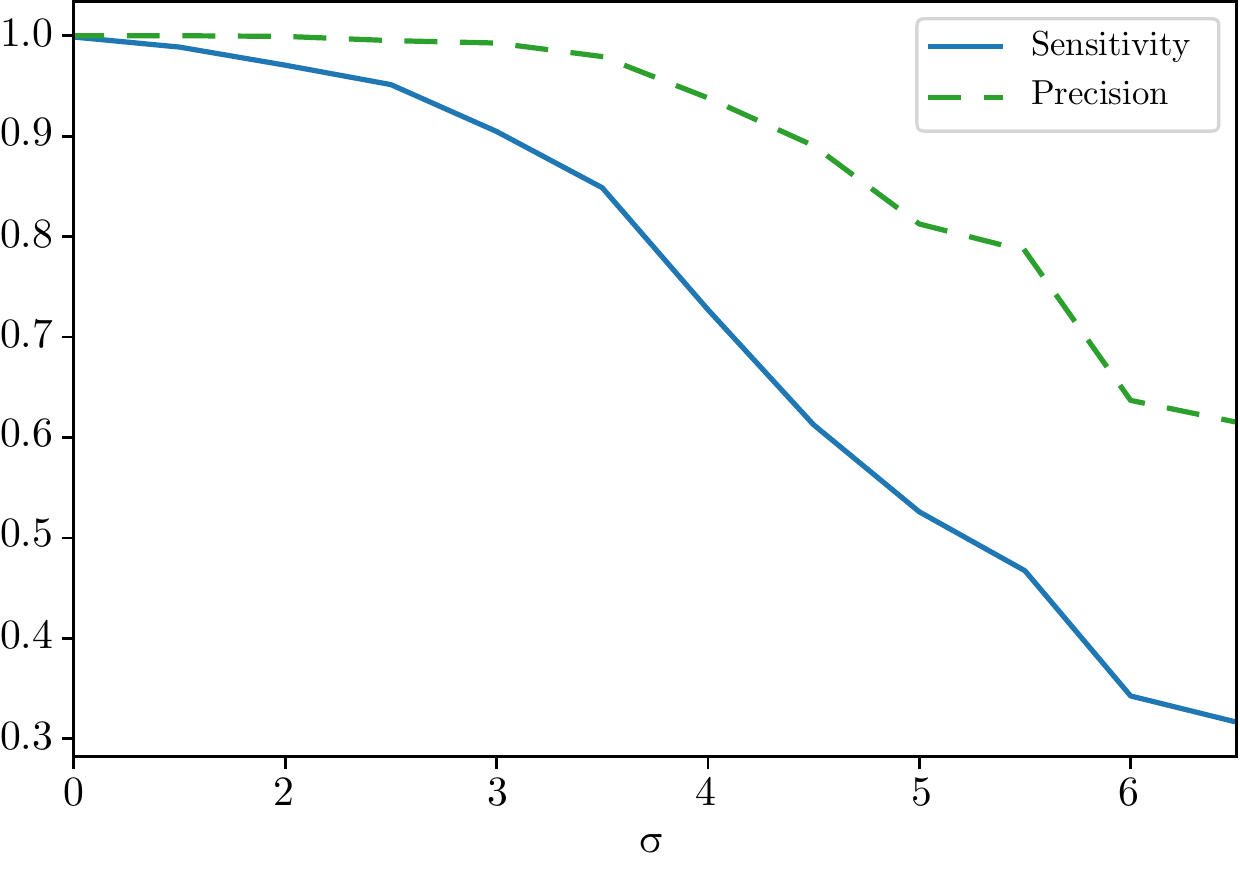}
        \caption{Simulated shift variations}
%        \label{}
    \end{subfigure}
	%\hspace{0.05cm}
	\begin{subfigure}[t]{0.45\linewidth}
        \includegraphics[width=\linewidth]{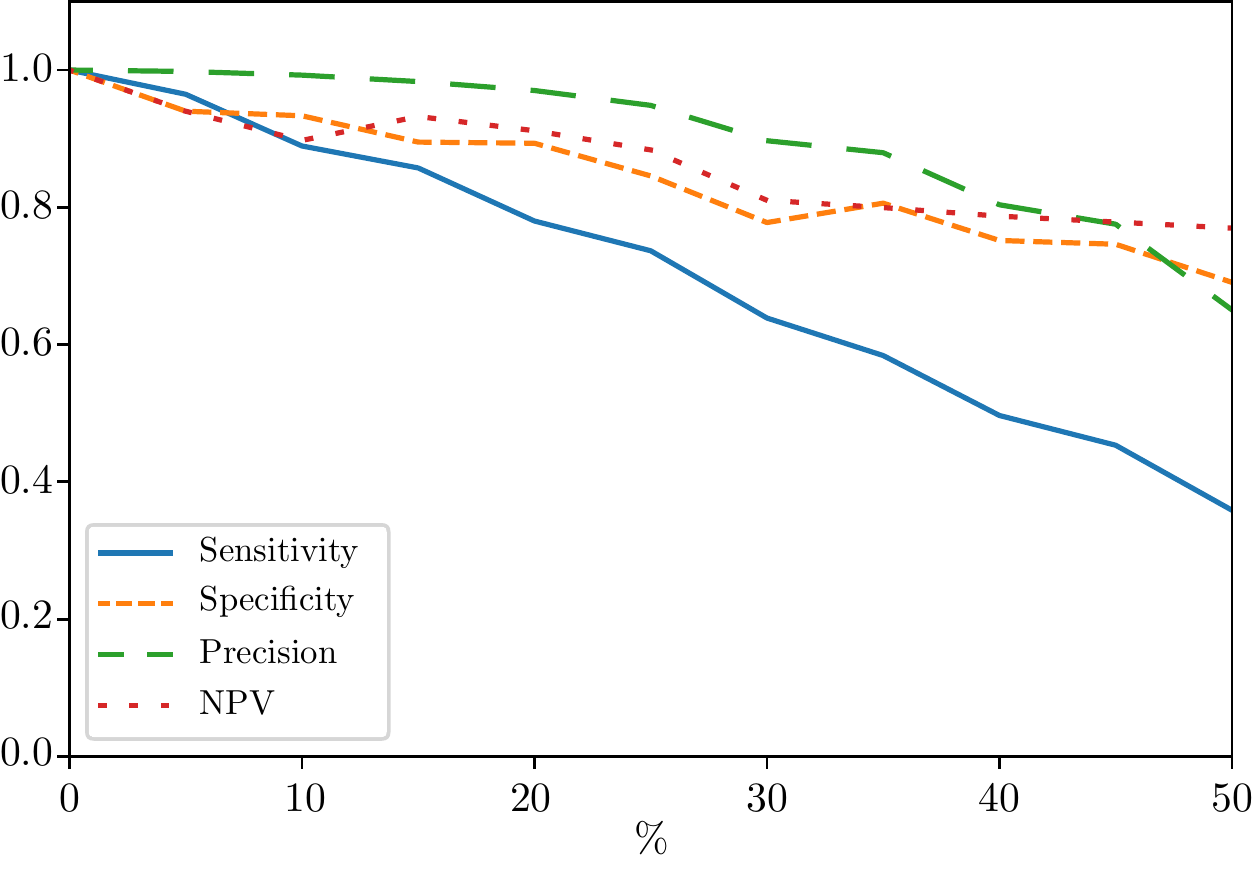}
        \caption{Simulated unpaired centroids}
%        \label{}
    \end{subfigure}
	\caption{The results of the matching algorithm on synthetic data with $d_{\text{match}} = 80$ and $N = 3$.}
	\label{fig:simulated_experiments}
\end{figure}

To evaluate the performance of the proposed algorithm to variations in the data, a simulated dataset was created. Fifty $300 \times 300$ image pairs were generated, each pair representing two consecutive slides. For each image in a pair, $30$ centroids were generated located at the same position in both images. The following two variations to the centroids were analysed.
\begin{description}[noitemsep,topsep=0pt,labelsep=0pt]
\item[Shift ] --- For each second image in a pair, the $x$ and $y$ position of each centroid was shifted by values drawn independently from a Gaussian distribution with $\mu = 0$ and $\sigma \in \{0\dots 11\}$ (that is $0$ to $3.6\%$ of the image size in each dimension).
\item[Unpaired ] ---
Spurious unpaired centroids were randomly added to each image in a pair. The number of centroids added to each image ranged from $0\%$ to $50\%$ of the initial number of centroids in the image. An addition of $50\%$ means that both images in a pair contains $45$ centroids but only $30$ should be matched. 

\end{description}

The Shift experiment was designed to evaluate the normal spatial variations of glomeruli slices in consecutive WSI whereas the Unpaired experiment evaluates the algorithm's behaviour to glomeruli appearance and disappearance between slides, and to errors during glomeruli segmentation.

Sensitivity ($S = \tfrac{\text{TP}}{\text{TP}+\text{FN}}$), and precision ($P = \tfrac{\text{TP}}{\text{TP}+\text{FP}}$) were measured during the Shift experiment. The values of TP, FP, and FN were measured in terms of associations, such that a TP is a correct association, an FP is an incorrect association, and an FN is when no association is made incorrectly. During the Unpaired experiment, Specificity ($SP = \tfrac{\text{TN}}{\text{TN}+\text{FP}}$) and Negative Predictive Value ($\text{NPV} = \tfrac{\text{TN}}{\text{TN}+\text{FN}}$) were also measured to account for the possibility of false positive---a centroid incorrectly associated to another---and true negative associations---unpaired centroids not associated with another correctly.

The average measure (over the $50$ repetitions of each setup) for each experiment is presented in Fig.\ \ref{fig:simulated_experiments}. These experiments show that the proposed algorithm is robust to shift and unpaired centroids. It is interesting to see that precision remains high with the increase of each parameter (shift and the number of added centroids) even though the specificity decreases more quickly. This means that the algorithm tends to avoid falsely associating glomeruli, which is a highly desirable behaviour when the goal is to extract statistical measures based on quantitative data extracted from image processing algorithms.

\subsection{Patients' Characteristics}
\label{sec:pc}

Patients’ characteristics are described in the following table, in which Banff assessments are made according to the 2013 Banff consensus \cite{Haas14}, Tx means Transplant, and NA means Not Available.
\\

{\begin{center}
\Rotatebox{270}{
\noindent\scriptsize{
\begin{tabular}{p{0.5cm} p{1cm} p{0.9cm} p{3.5cm} p{3.5cm} p{2.25cm}}
\hline
\hline
\multicolumn{1}{c}{\textbf{Gender}} & \centering{\textbf{Age (Years)}} & \centering{\textbf{Time Post Tx}} & \centering{\textbf{Reason for Tx Nephrectomy}} & \centering{\textbf{Pathological Observations}} & {\centering \textbf{Banff Classification}}
\\
\hline
\hline
Male & 63 & NA & Tumor in the renal graft & Clear cell renal cell carcinoma, severe atrophy and fibrosis\footnote{A renal carcinoma was present within the graft, the sample was taken as far as possible from the tumor, containing only non-malignant renal transplant tissue.} & Cat.\ 5, Grade III
\\
\hline
Male & 63 & 7 months & Recurrent severe viral
infections (Polyoma), loss of graft
function, renal inflammation
  & Combined cellular and humoral rejection, infarction, thrombotic blood vessel occlusion, severe atrophy and fibrosis
 & Cat.\ 2, Type III; Cat.\ 4, Type III; Cat.\ 5, Grade II
\\
\hline
Female & 77 & 5 years & Loss of graft function 3 years post tx, humoral rejection led to poor
general condition, renal inflammation
& End-stage renal graft with severe humoral rejection, severe atrophy and fibrosis & Cat.\ 2, Type III; Cat.\ 5, Grade III
\\
\hline
Male & 55 & 3 months & Non-functional symptomatic graft, renal inflammation & Cellular rejection, thrombs, moderate atrophy and fibrosis & Cat.\ 4, Type IIA; Cat.\ 5, Grade I
\\
\hline
\hline
\end{tabular}
}
}
\end{center}}
%\label{tbl:patient}
%\end{table}
\color{black}

\subsection{Staining Characteristics}
\label{sec:sc}

Immunohistochemstry staining was performed on consecutive 3µm thick paraffin section using an automated staining instrument (Ventana Benchmark Ultra) following the manufacturer’s recommendations, and using 3,3 diamino benzidine (DAB), or alkaline phosphatase (AP)/Fast-Red as chromogens. The primary antibodies used are described in the following table.
\\

%\Rotatebox{90}{
\noindent\scriptsize\centering{
\begin{tabular}{c c c c}
\hline
\hline
\textbf{Antigen} & \textbf{Clone} & \textbf{Provider} & \textbf{Order Number}
\\
\hline
%\hline
CD3 &	Polyclonal rabbit &	DAKO/Agilent &	A0452
\\
%\hline
CD206 &	5C11 (Monoclonal mouse) & 	BioRad &	MCA5552Z
\\
%\hline
MS4A4A &	Polyclonal rabbit &	Sigma &	HPA029323
\\
%\hline
CD163 &	MRQ-26(Monoclonal mouse) &	Cell Marque &	163M-14
\\
%\hline
CD68 & 	PG-M1 & 	DAKO/Agilent &	GA613
\\
\hline
%\hline
\end{tabular}
}
%}

\color{black}
\end{document}